\documentclass{article}

\usepackage[dblblindworkshop, final]{neurips_2026}
\workshoptitle{Neural Network Artifacts as a New Data Modality}

\usepackage[utf8]{inputenc} 
\usepackage[T1]{fontenc}    
\usepackage{hyperref}       
\usepackage{url}            
\usepackage{booktabs}       
\usepackage{amsfonts}       
\usepackage{nicefrac}       
\usepackage{microtype}      
\usepackage{xcolor}         

\usepackage{tikz}
\usetikzlibrary{positioning,fit,backgrounds,arrows.meta,calc,patterns}
\usepackage{amsmath}
\usepackage{subcaption}
\usepackage{wrapfig}
\usepackage[dvipsnames]{xcolor}
\usepackage{float}
\usepackage{pgfplots}
\usepgfplotslibrary{fillbetween}
\pgfplotsset{
    compat=1.18,
    cyclepanel/.style={
        axis lines=left,
        xmin=-0.4, xmax=10.4,
        xtick={0,1,2,3,4,5,6,7,8,9,10},
        xlabel={Cycle round},
        grid=both,
        grid style={line width=0.3pt, draw=black!15},
        tick align=outside,
        tick style={black!40, line width=0.4pt},
        label style={font=\small},
        tick label style={font=\footnotesize},
        title style={font=\small, yshift=2pt},
    },
    bandstyle/.style={
        teal!65!black,
        opacity=0.16
    },
    meanstyle/.style={
        teal!65!black,
        line width=1.1pt,
        mark=*,
        mark size=1.4pt,
        mark options={fill=teal!65!black, draw=teal!65!black}
    },
}

\title{Adapters as Data: The Latent Space of LoRAs}

\author{%
  Moritz Grävinghoff\textsuperscript{1,*} \\
  \And
  Konstantin Wehmeyer\textsuperscript{1} \\
  \And
  Aron Asefaw\textsuperscript{1} \\
  \AND
  Boris Knyazev\textsuperscript{2,3} \\
  \And
  Damian Borth\textsuperscript{1,3} \\
}

\begin{document}

\maketitle

{
\noindent
\centering
\textsuperscript{1}University of St.Gallen \hspace{5mm} 
\textsuperscript{2}Université de Montréal \hspace{5mm}
\textsuperscript{3}Weight Space Labs
\par
\vspace{10mm}
}

{\let\thefootnote\relax\footnotetext{\textsuperscript{*}Correspondence to: \texttt{moritz.graevinghoff@unisg.ch}. Code at: \url{https://github.com/HSG-AIML/Adapters_as_Data}}}

\begin{abstract}
LoRA adapters are usually treated as finished artifacts rather than as a data modality. Prior work has shown that LoRA adapters populate a low-dimensional manifold that a linear PCA basis already organizes well enough to support different operations such as sampling, editing, and inversion. We ask what a learned, nonlinear latent adds beyond this. We train a transformer autoencoder on the same population of LoRA identity adapters, mapping each into a latent space and compare every operation against the identical operation applied directly to the raw adapter weights. The latent preserves the capabilities already available in weight space, sampling diverse identities and editing individual attributes. Beyond preservation, we show that the latent is far more robust to displacement than weight space, degrading into distorted but still valid faces rather than texture. Fixed-dimensional weight-space representations cannot natively host adapters of more than one rank, so we extend our sequential autoencoder to hold mixed-rank adapters within one shared latent. This gives a path toward treating adapter weights of varying architecture as a single, general modality rather than one tied to a fixed rank, and points to a learned, nonlinear latent as the mechanism that could carry this generalization further, toward heterogeneous populations of adapters.
\end{abstract}

\section{Introduction}

Public repositories now host hundreds of thousands of LoRA adapters, yet each is still produced by its own training run and treated as a finished artifact. A network's weights are also points in a high-dimensional space, and once many networks exist these points form a broader distribution rather than existing in isolation. This premise underlies \emph{weight space learning}, which treats the parameters of neural networks as a data modality in their own right and learns representations that predict model properties directly from parameters~\citep{unterthiner2020predicting, schurholt2021selfsupervised, putterman2025gl} or synthesize new weights without training them from scratch~\citep{schurholt2022hyperrepresentations, schurholt2024sane, knyazev2023canwescale, wang2024neural,asefaw2026weightclip,hanna2026geosane}.

A narrower section of this literature has begun to ask the same question of Low-Rank Adaptation~(LoRA), now the default mechanism for adapting large foundation models such as Stable Diffusion~\citep{rombach2022ldm} to new identities, styles, and tasks. Since each adapter is still trained separately, composition of existing adapters remains a heuristic and unstable blend, and an adapter's behavior is hard to predict before generation. weights2weights (w2w)~\citep{dravid2024interpreting} was first to show that a large collection of identity-encoding LoRAs occupies a low-dimensional manifold that a simple PCA basis already organizes well enough to support sampling, editing, and inversion. We build directly on this setting, using the same base model, the same identity-encoding rank-1 adapters, and the same released dataset, but explore a different direction. Other methods generate LoRA weights outright rather than characterizing the space they come from, whether by pairing an autoencoder with a diffusion prior, mapping task prompts directly to weights, or predicting an adapter from a single image within a fixed subspace, and each is evaluated mainly by the quality of the adapter it produces~\citep{jin2024conditional, liang2025dragdrop, ruiz2024hyperdreambooth}. What remains open is whether the space a method samples from or restricts itself to is organized in a stronger sense than downstream performance alone can reveal. That would mean neighborhoods smooth under perturbation, operations that behave consistently rather than only within the convex hull of what was fit, and a representation flexible enough to host more than one fixed adapter architecture at once.

We study this question on a sample of w2w's population of roughly $65{,}000$ identity-encoding LoRA adapters, each fine-tuned on a shared frozen diffusion backbone to reproduce one synthetic face identity and annotated with binary CelebA attributes. Rank-1 adapters are our primary setting, matching w2w directly, and we later extend to rank-2 adapters as well. Using the SANE methodology~\citep{schurholt2024sane}, these weights are tokenized and an autoencoder is trained in a self-supervised manner to map them into a latent space $\mathcal{Z}$, with no contrastive objective, and no attribute conditioning. Our central question is whether $\mathcal{Z}$ is a \emph{structured} representation of the adapter population, rather than merely a compression of it. We probe this along three axes: (i) whether operations already known to work directly on raw adapter weights, sampling new identities and steering individual attributes, are preserved once routed through a learned, nonlinear latent; (ii) whether that latent is more robust to displacements than the raw weight space; and (iii) whether a single autoencoder can absorb adapters of more than one rank. Our contributions are as follows.

\begin{itemize}
    \item We introduce a methodology for tokenizing LoRA adapters into weight sequences and embedding them into a structured latent space with a self-supervised transformer autoencoder.
    \item We show that the resulting latent preserves the sampling and editing capabilities already present in the raw weight space, is substantially more robust to displacement, and remains stable under repeated encode-decode cycles.
    \item We extend the representation to multi-rank LoRA adapters via a padding scheme that lets a single autoencoder process adapters of different rank within one shared latent.
\end{itemize}

\section{Related Work}

\paragraph{Weight Space Learning.}
Representation learning over neural network weights spans both model analysis and weight generation. Regarding model analysis, work focuses on predicting model properties from weights \citep{unterthiner2020predicting, eilertsen2020classifying} or studies the structure of trained weights itself, motivating permutation- and equivariance-aware representations~\citep{navon2023equivariant, zhou2023permutation,kofinas2024graph,lim2024graph}. Closer to what we build on, a separate line of work treats populations of weights themselves as a data modality independent of any specific downstream task. \citet{schurholt2021selfsupervised} introduce self-supervised embeddings of model populations and subsequently show these can be used generatively, sampling unseen but performant weights for initialization, ensembling, and transfer learning~\citep{schurholt2022hyperrepresentations}. SANE~\citep{schurholt2024sane} scales this idea beyond small networks by tokenizing weight matrices into a sequence and processing it with a transformer autoencoder, which keeps the per-token dimensionality independent of overall model size, and is the architecture we adopt in this work. In parallel, weight generation has also been explored directly with diffusion and other generative priors~\citep{wang2024neural,wang2025scaling,soro2025diffusion}. A broader overview is provided in \citet{han2026wsl_survey,ettling2026weight}. Collectively, this body of work establishes that weight tensors can be treated as a learnable distribution with its own latent structure.

\paragraph{Attribute Editing and Adapters.}
Low-Rank Adaptation (LoRA)~\citep{hu2022lora} freezes a pretrained model and learns a low-rank update for selected weight matrices, typically at a single fixed rank shared across an entire collection of adapters, and is the adapter format we study throughout. FVAE-LoRA~\citep{kumar2025latent} further separates task-salient from residual information through variational factorization of layer activations during fine-tuning, whereas we learn a latent over adapter weights. Existing adapters can be combined post hoc, e.g.\ ZipLoRA~\citep{shah2024ziplora} merges an independently trained subject and style LoRA via a learned coefficient mask, but such methods treat each adapter as fixed and optimize a merging procedure rather than a shared latent. Attribute control is more commonly approached through a learned direction in an existing latent: Concept Sliders~\citep{gandikota2024sliders} train a dedicated LoRA per attribute on a frozen diffusion model, while InterFaceGAN~\citep{shen2020interfacegan}, GANSpace~\citep{harkonen2020ganspace}, and SeFa~\citep{shen2021closedform} fit, respectively, a linear SVM boundary, an unsupervised PCA direction, and a closed-form eigendecomposition to steer a pretrained GAN's latent. We apply the same two ideas, a fitted direction for editing and PCA for diversity, one level removed to a latent over adapter weights instead of images.

\paragraph{Modeling LoRA Populations.}
A separate line of work generates LoRA adapters directly from a population, rather than editing or combining existing ones one at a time. COND P-DIFF~\citep{jin2024conditional} adds a conditional diffusion prior over an autoencoded LoRA latent, and Drag-and-Drop LLMs~\citep{liang2025dragdrop} uses a hypernetwork to map task prompts directly to LoRA weight updates in a single forward pass, without any latent space. We instead embed adapters without any external conditioning signal, using only the weights themselves, so the same encoder can in principle be applied to any adapter, rather than requiring a paired task description or prompt. weights2weights~(w2w)~\citep{dravid2024interpreting} shows that a large collection of identity-encoding LoRAs already lies on a manifold that a simple PCA basis organizes well enough to support sampling, editing, and inversion, and is the direct baseline we build on and compare against throughout this paper. All existing methods are built around adapters of a single fixed rank and offer no way to mix ranks without being rebuilt from scratch. We address that limitation directly by extending a learned latent to hold adapters of more than one rank in a single shared representation.

\section{Approach}
\label{sec:approach}
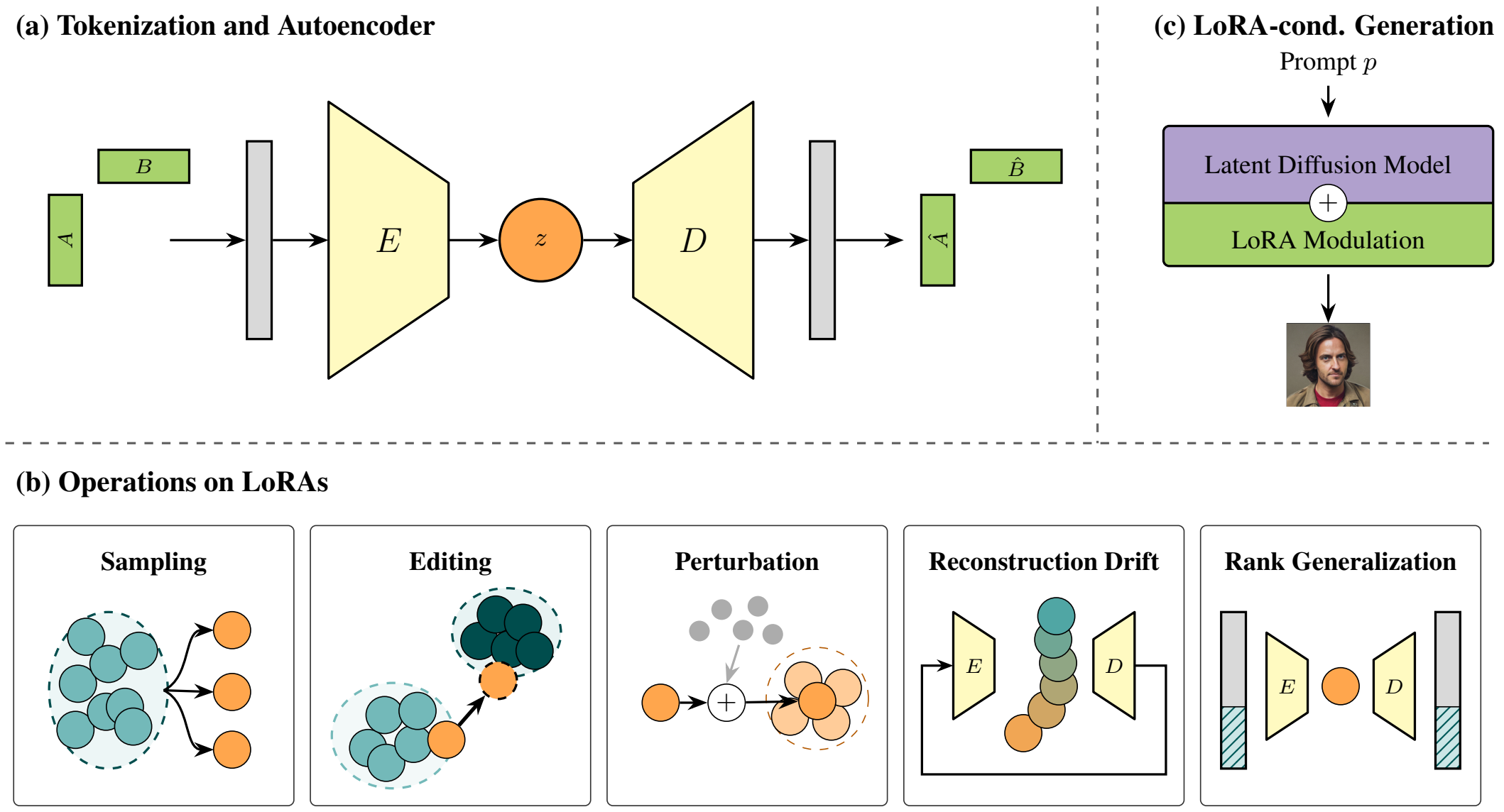
\begin{figure*}[t]
\centering

\resizebox{\textwidth}{!}{%
\begin{tikzpicture}[
    font=\small,
    >={Stealth[length=2.4mm]},
    flow/.style={->, thick},
    matrixA/.style={
        draw,
        thick,
        fill=LimeGreen!75,
        minimum width=11mm,
        minimum height=4mm,
        inner sep=1pt,
        font=\scriptsize
    },
    matrixB/.style={
        draw,
        thick,
        fill=LimeGreen!75,
        minimum width=4mm,
        minimum height=11mm,
        inner sep=1pt,
        font=\scriptsize
    },
    tokenvec/.style={
        draw,
        thick,
        fill=gray!30,
        minimum width=3mm,
        minimum height=24mm,
        inner sep=0pt
    },
    aeBlock/.style={
        draw,
        thick,
        fill=yellow!30,
        line join=round
    },
    latent/.style={
        draw,
        circle,
        thick,
        fill=orange!70,
        minimum size=10mm
    },
    exbox/.style={
        draw=black!80,
        thin,
        rounded corners=2pt,
        minimum width=34mm,
        minimum height=34mm,
        inner sep=0pt
    }
]

\node[
    font=\bfseries,
    anchor=west
] at (0,2.25) {(a) Tokenization and Autoencoder};
\begin{scope}[
    scale=1.12,
    yshift=-3mm,
]
\node[matrixB] (inA) at (0.65,0)
    {\rotatebox{90}{$A$}};
\node[matrixA] (inB) at (1.5,0.8)
    {$B$};

\node[tokenvec] (tokensIn) at (2.75,0) {};

\draw[
    flow,
    shorten <=2mm
] ([xshift=-0.4cm,yshift=-0.8cm]inB.east) -- (tokensIn.west);

\coordinate (encL) at (3.5,0);
\coordinate (encR) at (4.8,0);

\draw[aeBlock]
    ($(encL)+(0,1.5)$) --
    ($(encR)+(0,0.62)$) --
    ($(encR)+(0,-0.62)$) --
    ($(encL)+(0,-1.5)$) --
    cycle;

\node[font=\large\itshape]
    at ($(encL)!0.5!(encR)$) {$E$};

\draw[flow] (tokensIn.east) -- (encL);

\node[latent] (z) at (5.8,0) {$z$};

\draw[flow] (encR) -- (z.west);

\coordinate (decL) at (6.8,0);
\coordinate (decR) at (8.1,0);

\draw[aeBlock]
    ($(decL)+(0,0.62)$) --
    ($(decR)+(0,1.5)$) --
    ($(decR)+(0,-1.5)$) --
    ($(decL)+(0,-0.62)$) --
    cycle;

\node[font=\large\itshape]
    at ($(decL)!0.5!(decR)$) {$D$};

\draw[flow] (z.east) -- (decL);

\node[tokenvec] (tokensOut) at (8.85,0) {};

\draw[flow] (decR) -- (tokensOut.west);

\node[matrixB] (outA) at (10.1,0)
    {\rotatebox{90}{$\hat A$}};
\node[matrixA] (outB) at (10.95,0.8)
    {$\hat B$};

\draw[
    flow,
    shorten >=2mm
] (tokensOut.east) -- (outA.west);
\end{scope}


\draw[
    dashed,
    thick,
    black!60
] (13.25,2.5) -- (13.25,-2.8);


\node[
    font=\bfseries
] at (16,2.25) {(c) LoRA-cond. Generation};
\begin{scope}[
    xshift=5mm,
    yshift=-2mm,
]
\coordinate (modelSW) at (13.55,-0.45);
\coordinate (modelNE) at (17.55,1.25);

\coordinate (splitL) at
    ($(modelSW)!0.45!(13.55,1.25)$);

\coordinate (splitR) at
    ($(17.55,-0.45)!0.45!(modelNE)$);

\coordinate (splitCenter) at
    ($(splitL)!0.5!(splitR)$);

\begin{scope}
    \clip[rounded corners=2pt]
        (modelSW) rectangle (modelNE);

    \fill[LimeGreen!75]
        (modelSW) rectangle (splitR);

    \fill[fill=RoyalPurple!40]
        (splitL) rectangle (modelNE);
\end{scope}

\draw[
    thick,
    rounded corners=2pt
] (modelSW) rectangle (modelNE);

\draw[thick] (splitL) -- (splitR);

\node[align=center] at (15.55,0.78)
    {Latent Diffusion Model};

\node[align=center] at (15.55,-0.13)
    {LoRA Modulation};

\node[
    circle,
    draw,
    thin,
    fill=white,
    inner sep=1.2pt
] at (splitCenter) {$+$};

\node (prompt) at (15.55,2.00)
    {Prompt $p$};

\draw[flow]
    (prompt.south) -- (15.55,1.35);

\node[
    draw,
    inner sep=0pt
] (generatedImage) at (15.55,-1.65) {
    \includegraphics[
        width=10mm,
        height=10mm,
        keepaspectratio
    ]{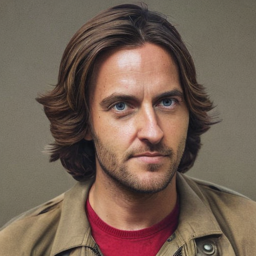}
};

\draw[flow]
    (15.55,-0.55) -- (generatedImage.north);
\end{scope}


\draw[
    dashed,
    thick,
    black!60
] (0,-2.8) -- (18,-2.8);


\node[
    font=\bfseries,
    anchor=west
] at (0,-3.3) {(b) Operations on LoRAs};

\node[exbox] (component1) at (1.8,-5.5) {};
\begin{scope}[
    shift={(component1.center)},
    populationdot/.style={
        circle,
        draw=black,
        thin,
        fill=teal!55,
        minimum size=4.5mm,
        inner sep=0pt
    },
    sampledot/.style={
        circle,
        draw=black,
        thin,
        fill=orange!70,
        minimum size=4.5mm,
        inner sep=0pt
    }
]

\node[
    font=\small\bfseries,
    anchor=north
] at (0,1.5) {Sampling};

\fill[
    teal!12,
    opacity=0.55
] (-0.55,-0.30) ellipse (0.72 and 0.95);

\draw[
    dashed,
    thick,
    teal!60!black
] (-0.55,-0.30) ellipse (0.72 and 0.95);

\coordinate (sampleSource) at (0.12,-0.30);

\draw[->,thick]
    (sampleSource) to[out=25,in=180] (0.73,0.43);

\draw[->,thick]
    (sampleSource) -- (0.73,-0.32);

\draw[->,thick]
    (sampleSource) to[out=-25,in=180] (0.73,-1.02);

\foreach \x/\y in {
    -0.95/-0.78,
    -0.92/-0.22,
    -0.82/0.35,
    -0.58/-0.56,
    -0.55/0.02,
    -0.35/-0.50,
    -0.25/-0.74,
    -0.18/0.18
} {
    \node[populationdot] at (\x,\y) {};
}

\node[sampledot] at (0.95,0.43) {};
\node[sampledot] at (0.95,-0.32) {};
\node[sampledot] at (0.95,-1.02) {};

\end{scope}

\node[exbox] (component2) at (5.4,-5.5) {};

\begin{scope}[
    shift={(component2.center)},
    lightcluster/.style={
        circle,
        draw=black,
        thin,
        fill=teal!55,
        minimum size=4.5mm,
        inner sep=0pt
    },
    darkcluster/.style={
        circle,
        draw=black,
        thin,
        fill=teal!60!black,
        minimum size=4.5mm,
        inner sep=0pt
    },
    zdot/.style={
        circle,
        draw=black,
        thin,
        fill=orange!70,
        minimum size=4.5mm,
        inner sep=0pt
    },
    zpdot/.style={
        circle,
        draw=black,
        thick,
        dashed,
        fill=orange!70,
        minimum size=4.5mm,
        inner sep=0pt
    }
]

\node[
    font=\small\bfseries,
    anchor=north
] at (0,1.5) {Editing};

\fill[
    teal!10,
    opacity=0.45
] (-0.70,-0.85) ellipse (0.74 and 0.64);

\draw[
    dashed,
    thick,
    teal!55
] (-0.70,-0.85) ellipse (0.74 and 0.64);

\fill[
    teal!20,
    opacity=0.45
] (0.68,0.40) ellipse (0.66 and 0.54);

\draw[
    dashed,
    thick,
    teal!60!black
] (0.68,0.40) ellipse (0.66 and 0.54);

\foreach \x/\y in {
    -1.00/-1.0,
    -0.78/-1.18,
    -0.80/-0.60,
    -0.45/-0.95,
    -0.40/-0.55
} {
    \node[lightcluster] at (\x,\y) {};
}

\foreach \x/\y in {
    0.35/0.30,
    0.55/0.62,
    0.72/0.20,
    0.88/0.52,
    1.02/0.18
} {
    \node[darkcluster] at (\x,\y) {};
}

\node[zdot]  (editZ)  at (-0.05,-0.90) {};
\node[zpdot] (editZp) at ( 0.58,-0.18) {};

\draw[
    ->,
    line width=1.3pt,
    black,
] (editZ) -- (editZp);

\end{scope}

\node[exbox] (component3) at (9,-5.5) {};

\begin{scope}[
    shift={(component3.center)},
    latentdot/.style={
        circle,
        draw=black,
        thin,
        fill=orange!70,
        minimum size=4.5mm,
        inner sep=0pt
    },
    noisylatent/.style={
        circle,
        draw=black,
        thin,
        fill=orange!40,
        minimum size=4.5mm,
        inner sep=0pt
    },
    noiseparticle/.style={
        circle,
        fill=gray!65,
        minimum size=2.5mm,
        inner sep=0pt
    }
]

\node[
    font=\small\bfseries,
    anchor=north
] at (0,1.5) {Perturbation};

\node[latentdot] (cleanZ) at (-1.05,-0.45) {};

\node[
    circle,
    draw,
    thin,
    fill=white,
    inner sep=1.2pt
] (noiseAdd) at (-0.25,-0.45) {$+$};

\draw[->,thick] (cleanZ) -- (noiseAdd);

\foreach \x/\y in {
    -0.58/0.42,
    -0.31/0.68,
    -0.05/0.43,
     0.13/0.72,
     0.31/0.38
} {
    \node[noiseparticle] at (\x,\y) {};
}

\draw[
    ->,
    thick,
    gray!65
] (-0.10,0.25) -- (noiseAdd.north);

\draw[
    dashed,
    thin,
    orange!70!black
] (0.85,-0.40) circle (0.62);

\node[noisylatent] at (0.52,-0.62) {};
\node[noisylatent] at (0.68,-0.18) {};
\node[noisylatent] at (1.02,-0.68) {};
\node[noisylatent] at (1.16,-0.27) {};

\node[latentdot] (perturbedZ) at (0.85,-0.43) {};

\draw[->,thick] (noiseAdd) -- (perturbedZ);

\end{scope}

\node[exbox] (component4) at (12.6,-5.5) {};

\begin{scope}[
    shift={(component4.center)},
    miniAE/.style={
        draw,
        thick,
        fill=yellow!30,
        line join=round
    },
    feedback/.style={
        -{Stealth[length=2mm]},
        thick,
        black
    }
]

\node[
    font=\small\bfseries,
    anchor=north
] at (0,1.5) {Reconstruction Drift};

\coordinate (miniEncL) at (-1.1,0);
\coordinate (miniEncR) at (-0.6,0);

\draw[miniAE]
    ($(miniEncL)+(0,0.65)$) --
    ($(miniEncR)+(0,0.32)$) --
    ($(miniEncR)+(0,-0.32)$) --
    ($(miniEncL)+(0,-0.65)$) --
    cycle;

\node[font=\scriptsize\itshape]
    at ($(miniEncL)!0.5!(miniEncR)$) {$E$};

\coordinate (miniDecL) at (0.6,0);
\coordinate (miniDecR) at (1.1,0);

\draw[miniAE]
    ($(miniDecL)+(0,0.32)$) --
    ($(miniDecR)+(0,0.65)$) --
    ($(miniDecR)+(0,-0.65)$) --
    ($(miniDecL)+(0,-0.32)$) --
    cycle;

\node[font=\scriptsize\itshape]
    at ($(miniDecL)!0.5!(miniDecR)$) {$D$};

\foreach \i in {0,1,2,3,4,5} {
    \pgfmathsetmacro{\t}{\i/5}
    \pgfmathsetmacro{\px}{
        -0.25 + 0.6*\t + 0.2*sin(270*\t)
    }
    \pgfmathsetmacro{\py}{
        -0.82 + 1.42*\t
    }
    \pgfmathtruncatemacro{\shade}{8 + 18*\i}

    \node[
        circle,
        draw=black,
        thin,
        fill=teal!\shade!orange!70,
        minimum size=4.5mm,
        inner sep=0pt
    ] (miniDrift\i) at (\px,\py) {};
}

\draw[feedback]
    (miniDecR) --
    (1.48,0) --
    (1.48,-1.32) --
    (-1.48,-1.32) --
    (-1.48,0) --
    (miniEncL);

\end{scope}

\node[exbox] (component5) at (16.2,-5.5) {};

\begin{scope}[
    shift={(component5.center)},
    miniAE/.style={
        draw,
        thick,
        fill=yellow!30,
        line join=round
    },
    ranklatent/.style={
        circle,
        draw=black,
        thin,
        fill=orange!70,
        minimum size=4.5mm,
        inner sep=0pt
    }
]

\node[
    font=\small\bfseries,
    anchor=north
] at (0,1.5) {Rank Generalization};


\coordinate (rankTokenIn) at (-1.35,0);

\fill[gray!30]
    ($(rankTokenIn)+(-0.10,-0.5)$)
    rectangle
    ($(rankTokenIn)+(0.20,0.65)$);

\fill[teal!20]
    ($(rankTokenIn)+(-0.10,-1.25)$)
    rectangle
    ($(rankTokenIn)+(0.20,-0.5)$);

\path[
    pattern=north east lines,
    pattern color=teal!70!black
]
    ($(rankTokenIn)+(-0.10,-1.25)$)
    rectangle
    ($(rankTokenIn)+(0.20,-0.5)$);

\draw[thick]
    ($(rankTokenIn)+(-0.10,-1.25)$)
    rectangle
    ($(rankTokenIn)+(0.20,0.65)$);

\draw[thin]
    ($(rankTokenIn)+(-0.10,-0.5)$) --
    ($(rankTokenIn)+(0.20,-0.5)$);


\coordinate (rankEncL) at (-0.9,-0.25);
\coordinate (rankEncR) at (-0.4,-0.25);

\draw[miniAE]
    ($(rankEncL)+(0,0.65)$) --
    ($(rankEncR)+(0,0.32)$) --
    ($(rankEncR)+(0,-0.32)$) --
    ($(rankEncL)+(0,-0.65)$) --
    cycle;

\node[font=\scriptsize\itshape]
    at ($(rankEncL)!0.5!(rankEncR)$) {$E$};

\node[ranklatent] at (0,-0.25) {};


\coordinate (rankDecL) at (0.4,-0.25);
\coordinate (rankDecR) at (0.9,-0.25);

\draw[miniAE]
    ($(rankDecL)+(0,0.32)$) --
    ($(rankDecR)+(0,0.65)$) --
    ($(rankDecR)+(0,-0.65)$) --
    ($(rankDecL)+(0,-0.32)$) --
    cycle;

\node[font=\scriptsize\itshape]
    at ($(rankDecL)!0.5!(rankDecR)$) {$D$};


\coordinate (rankTokenOut) at (1.35,0);

\fill[gray!30]
    ($(rankTokenOut)+(-0.20,-0.5)$)
    rectangle
    ($(rankTokenOut)+(0.10,0.65)$);

\fill[teal!20]
    ($(rankTokenOut)+(-0.20,-1.25)$)
    rectangle
    ($(rankTokenOut)+(0.10,-0.5)$);

\path[
    pattern=north east lines,
    pattern color=teal!70!black
]
    ($(rankTokenOut)+(-0.20,-1.25)$)
    rectangle
    ($(rankTokenOut)+(0.10,-0.5)$);

\draw[thick]
    ($(rankTokenOut)+(-0.20,-1.25)$)
    rectangle
    ($(rankTokenOut)+(0.10,0.65)$);

\draw[thin]
    ($(rankTokenOut)+(-0.20,-0.5)$) --
    ($(rankTokenOut)+(0.10,-0.5)$);

\end{scope}

\end{tikzpicture}%
}

\caption{
Overview of our approach. (a) The LoRA matrices $A$ and $B$ are tokenized and encoded by a transformer encoder $E$ into a latent $z$, which a decoder $D$ reconstructs back into an adapter. (b) We study five operations on this latent: sampling new identities, editing individual attributes, perturbing an adapter by a controlled amount, testing stability under repeated encode-decode cycles, and extending the latent to adapters of more than one rank. (c) The reconstructed or manipulated adapter modulates a frozen latent diffusion model to generate an image conditioned on a prompt $p$.
}
\label{fig:approach}
\end{figure*}

We take a single LoRA identity adapter $\theta = (A, B)$, trained on top of a frozen base model, and embed it into a structured latent space from which new adapters can be sampled and edited. Figure~\ref{fig:approach} shows the full path. Following the SANE scheme~\citep{schurholt2024sane}, the adapter matrices are reshaped and sliced row-wise into a sequence of fixed-size tokens, yielding a $[312 \times 320]$ token sequence for the rank-1 adapters that form our primary setting, together with a positional encoding and a mask marking real weights against padding. A GPT-style, non-causal transformer encoder $E$ maps this sequence to a latent $z = E(\theta) \in \mathbb{R}^{312 \times 256}$, preserving the token axis, and a symmetric decoder $D$ reconstructs the tokenized weights from it, trained with a masked reconstruction objective. Once an adapter is represented as a latent $z$, we study a range of operations on that space. We sample new latents that decode to novel but coherent adapters and steer a latent along learned directions to edit specific attributes (Section~\ref{sec:ex-part1}). We probe the geometry of that space under perturbation and test its stability under repeated encode-decode cycles (Section~\ref{sec:ex-part2}). We extend the representation itself to host adapters of more than one rank within a single shared latent (Section~\ref{sec:ex-part3}). In every case the manipulation happens on $z$ and is realized by decoding and de-tokenizing the latent into an adapter $\hat{\theta}$, applying it to the frozen base model, and rendering it to an image under a fixed prompt and seed, so that differences between renders are attributable to the adapter rather than to the rendering process. When applicable, every operation we apply to $z$ is additionally applied directly to $\theta$, bypassing the autoencoder entirely. This weight-space control is what lets us attribute any structure we find to the latent representation rather than to the adapter population itself. Architecture, training, and tokenization details, including the alternative scheme used for the mixed-rank setting (Section~\ref{sec:ex-part3}), are given in Appendix~\ref{app:implementation}.

\section{Experimental Setup}

\subsection{Data}
\label{sec:data}

Our adapters are LoRA fine-tunes of a single, shared, frozen base diffusion model, Realistic Vision v5.1\footnote{\url{https://huggingface.co/stablediffusionapi/realistic-vision-v51}}, a community fine-tune of Stable Diffusion~1.5~\citep{rombach2022ldm} optimized for photorealistic portraiture. Sharing one base model is what makes it meaningful to compare, interpolate, and embed different adapters, since every $\theta$ modifies the same underlying weights. Each adapter is trained with a DreamBooth-style objective~\citep{ruiz2023dreambooth} to reproduce one synthetic identity under a fixed identifier token. For our primary, rank-1 setting we follow~\citet{dravid2024interpreting}, whose released dataset and binary CelebA~\citep{liu2015celeba} attribute labels we use throughout. The dataset comprises $64{,}974$ identity adapters labeled along $40$ CelebA attributes with widely varying prevalence, from near-balanced (\emph{Male}, $48.3\%$) to rare (\emph{Bald}, $2.5\%$). One attribute, \emph{Wearing Hat}, is constant at $0\%$ across the dataset and is excluded from any attribute-conditioned experiment, leaving $39$ variable attributes. From this pool we subsample $8{,}192$ adapters, split into train, validation, and test, giving $\sim\!5{,}734$ training adapters. The full set of released adapters corresponds to only $9{,}256$ distinct underlying identities (median $6$ adapters each), and our splits are drawn per adapter rather than per identity, so near-duplicate fits of the same person can appear across different splits. For the rank-2 setting used in our mixed-rank experiments, no equivalent released population exists, so we train a smaller pool of rank-2 adapters ourselves, following the same base model, DreamBooth objective, and identifier-token convention as the rank-1 dataset. Details are given in Appendix~\ref{app:implementation}.

\subsection{Evaluation}
\label{sec:method-eval}

Most of our experiments ask qualitative questions, e.g.\ whether a sampled latent decodes to a genuinely new person, or whether an edit moved one attribute while leaving the rest of the identity unchanged, and a quantitative score can only approximate this. We nonetheless report two scores where they are informative. For weight-space fidelity, we use the coefficient of determination $R^2(\theta, \hat\theta)$ between an adapter and its reconstruction, and for image-space fidelity, we compute the LPIPS distance~\citep{zhang2018unreasonable} between the renders of an original and a reconstructed or manipulated adapter. Neither is a reliable proxy for identity preservation on its own and aggregating either can hide exactly the qualitative distinctions our experiments turn on, such as an adapter degrading into texture versus into a different but valid face. We therefore treat rendered figures as the primary evidence throughout, and use $R^2$ and LPIPS to establish trends across many checkpoints or identities, and to surface differences too small to see reliably.  Across $32$ checkpoint--identity pairs, the two scores are in fact tightly coupled (Pearson and Spearman $\rho = -0.90$, $p < 10^{-11}$, also holding under SSIM~\citep{wang2004ssim} and pixel-wise MSE), confirming that weight-space $R^2$ and perceptual quality track each other closely.

\section{The Latent Space Preserves Properties of the Weight Space}
\label{sec:ex-part1}

In this section we test whether operations already known to work directly on raw adapter weights~\citep{dravid2024interpreting} also work when routed through a learned, nonlinear latent trained with no label supervision. We examine two such operations here, sampling new identities and editing existing ones, and defer a third, arithmetic and interpolation between two identities, to Appendix~\ref{app:arithmetic-interpolation}.

\subsection{Sampling}
\label{sec:sampling}

\emph{Can a latent code drawn without reference to any training adapter be decoded into a new, coherent identity, and if so, are the draws diverse rather than collapsing onto a single average?}

\paragraph{Setup.} The simplest scheme, \emph{naive sampling}, fits an independent Gaussian to each latent coordinate, using the per-dimension mean and standard deviation of the encoded training latents, and draws each coordinate independently from its own marginal. A sample is drawn as a full $[312 \times 256]$ latent sequence. Because independent per-dimension sampling discards the covariance structure of the latent, the correlations between coordinates that encode which combinations of features co-occur among real identities, we also sample in the principal-component basis of the encoded training latents. A PCA is fitted to the flattened training latents, the top $k$ components are retained, and a sample is drawn as
\[
    z_{\text{sample}} = \bar{z} + \eta \sum_{j=1}^{k} \xi_j\,\sqrt{\lambda_j}\,u_j,
    \qquad \xi_j \sim \mathcal{N}(0,1),
\]
where $\bar{z}$ is the mean latent, $u_j$ and $\lambda_j$ are the $j$-th principal component and its variance, and $\eta$ is a \emph{diversity factor} that trades realism against variety: $\eta=1$ reproduces the natural spread of the population, larger $\eta$ pushes samples further from the mean.

\paragraph{Results.}
\begin{wrapfigure}[16]{r}{0.65\textwidth}
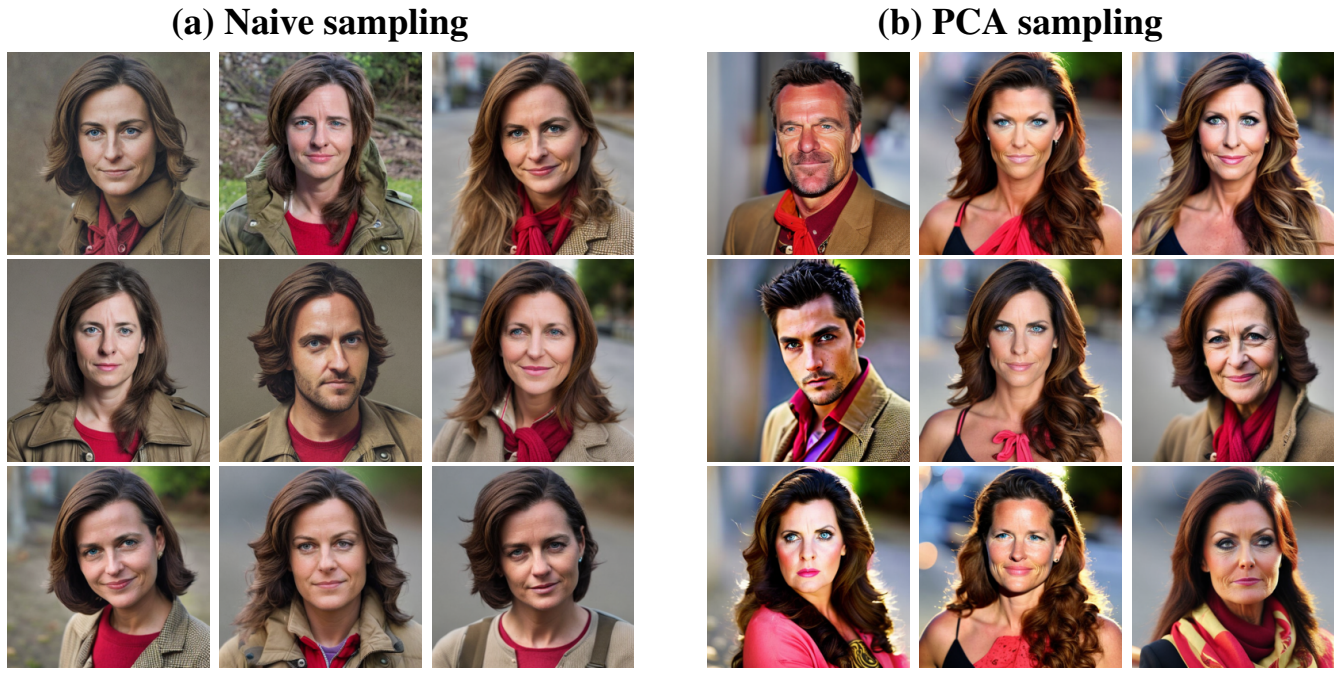

\centering
\vspace{-10pt}
\newcommand{\sampleimg}[1]{%
    \includegraphics[width=0.30\linewidth]{#1}%
}
\begin{minipage}[t]{0.49\linewidth}
\centering
\scriptsize\textbf{(a) Naive sampling}\par\vspace{1pt}
{\setlength{\tabcolsep}{1pt}
\renewcommand{\arraystretch}{0.4}
\begin{tabular}{@{}ccc@{}}
    \sampleimg{figures/04\_sampling\_naive/sample\_0\_x\_hat.png} &
    \sampleimg{figures/04\_sampling\_naive/sample\_1\_x\_hat.png} &
    \sampleimg{figures/04\_sampling\_naive/sample\_2\_x\_hat.png} \\
    \sampleimg{figures/04\_sampling\_naive/sample\_3\_x\_hat.png} &
    \sampleimg{figures/04\_sampling\_naive/sample\_4\_x\_hat.png} &
    \sampleimg{figures/04\_sampling\_naive/sample\_5\_x\_hat.png} \\
    \sampleimg{figures/04\_sampling\_naive/sample\_6\_x\_hat.png} &
    \sampleimg{figures/04\_sampling\_naive/sample\_7\_x\_hat.png} &
    \sampleimg{figures/04\_sampling\_naive/sample\_8\_x\_hat.png} \\
\end{tabular}}
\end{minipage}\hfill
\begin{minipage}[t]{0.49\linewidth}
\centering
\scriptsize\textbf{(b) PCA sampling}\par\vspace{1pt}

{\setlength{\tabcolsep}{1pt}
\renewcommand{\arraystretch}{0.4}
\begin{tabular}{@{}ccc@{}}
    \sampleimg{figures/04\_sampling\_PCA/sample\_0\_x\_hat.png} &
    \sampleimg{figures/04\_sampling\_PCA/sample\_1\_x\_hat.png} &
    \sampleimg{figures/04\_sampling\_PCA/sample\_2\_x\_hat.png} \\
    \sampleimg{figures/04\_sampling\_PCA/sample\_3\_x\_hat.png} &
    \sampleimg{figures/04\_sampling\_PCA/sample\_4\_x\_hat.png} &
    \sampleimg{figures/04\_sampling\_PCA/sample\_5\_x\_hat.png} \\
    \sampleimg{figures/04\_sampling\_PCA/sample\_6\_x\_hat.png} &
    \sampleimg{figures/04\_sampling\_PCA/sample\_7\_x\_hat.png} &
    \sampleimg{figures/04\_sampling\_PCA/sample\_8\_x\_hat.png} \\
\end{tabular}}
\end{minipage}

\caption{(a) Naive sampling collapses toward average-looking identities. (b) PCA sampling with $\eta=1$ and $k=256$ preserves substantially greater diversity.}
\label{fig:sampling-comparison}
\vspace{-6pt}
\end{wrapfigure}
Naive sampling produces valid, well-formed faces, but nearly the same face every time (Figure~\ref{fig:sampling-comparison}, left). Decoded identities share pose, coloring, and features to the point of looking like minor variations of one person, confirming that samples decode to coherent faces but collapsing toward an average identity rather than generating diversity. Sampling in the principal-component basis recovers the missing variety (Figure~\ref{fig:sampling-comparison}, right). With $k=256$ components and $\eta=1$, decoded samples are clearly distinct identities, differing in gender, age, hairstyle, and coloring, while remaining coherent and realistic. So many components are needed because the latent's variance is spread broadly rather than concentrated in a few directions. The leading component alone explains only $3.1\%$ of it, the top $10$ only $11.9\%$, and even the top $256$ together capture just $40.9\%$.

\paragraph{Relation to w2w.} weights2weights~\citep{dravid2024interpreting} follows the same recipe directly on raw adapter weights, fitting a PCA basis over the flattened LoRA updates and sampling coefficients along the leading components to obtain new identities. That we need the identical construction here is consistent with our autoencoder never being pushed to disentangle. Since it was trained with a reconstruction objective alone and no prior encouraging independent, unit-scale dimensions, there is no reason to expect the latent's coordinate axes to align with the population's true directions of variation, and naive per-dimension sampling requires exactly that alignment to work.

\newpage
\subsection{Editing}

\emph{Given one identity, can a single named attribute be changed continuously while the rest of the identity is left intact?}

\paragraph{Setup.} We estimate an attribute direction $v$ by ridge regression, where training latents are flattened into $Z \in \mathbb{R}^{N \times D}$ ($N = 5{,}734$, $D = 79{,}872$), CelebA labels are mapped to $y \in \{0,1\}^N$, and $v$ is the ridge solution $v = \arg\min_w \lVert Z_c w - y_c \rVert_2^2 + \lambda \lVert w \rVert_2^2$ ($\lambda = 1$) on centered data, solved in its dual form since $D \gg N$. This serves a similar purpose to the implicit shrinkage w2w~\citep{dravid2024interpreting} gets from truncating its PCA basis, though ridge shrinks every direction smoothly rather than discarding low-variance ones outright. An identity is edited as $z' = z + \gamma\,v$ (with $\gamma = 0$ giving back the reconstruction), where $\gamma$ is expressed as a fraction of $\lVert z \rVert$ so that edit strength is comparable across identities and directions. Before rendering any sweep, we also score each direction $v$ by the held-out AUC of adapters ranked by their projection onto $v$, the probability that a randomly chosen positive outranks a randomly chosen negative.

\begin{figure}[t]
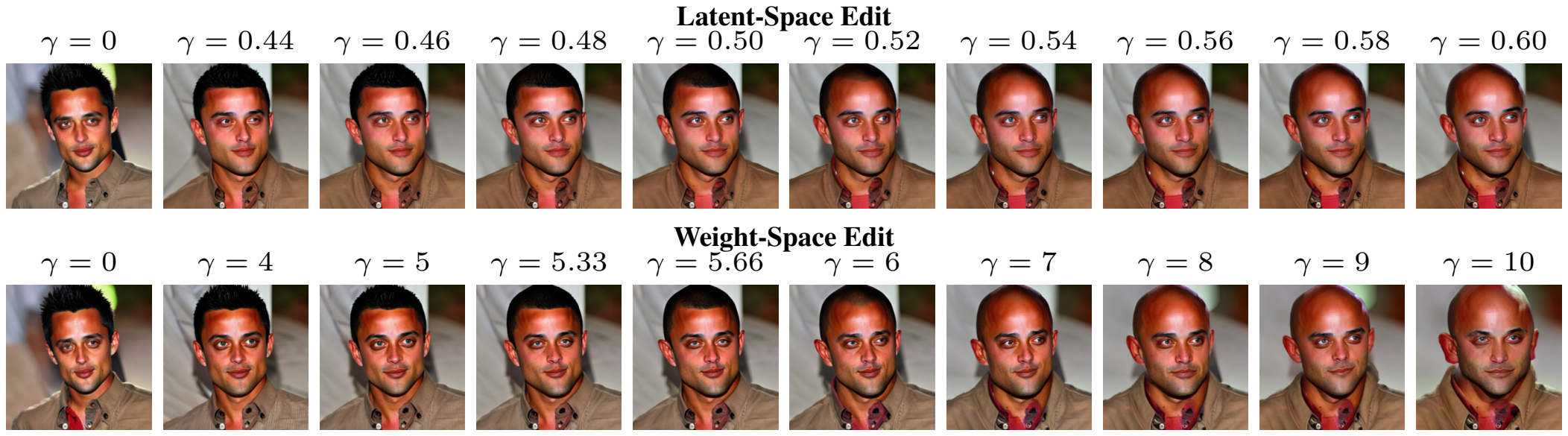

\centering

\setlength{\tabcolsep}{1.5pt}
\renewcommand{\arraystretch}{0.5}

\newcommand{\editimg}[1]{%
    \includegraphics[width=0.092\linewidth,height=0.092\linewidth]{#1}%
}

\newcommand{\latenteditrow}[1]{%
    \editimg{#1/sample\_0\_x\_hat.png} &
    \editimg{#1/sample\_3\_x\_hat.png} &
    \editimg{#1/sample\_4\_x\_hat.png} &
    \editimg{#1/sample\_5\_x\_hat.png} &
    \editimg{#1/sample\_6\_x\_hat.png} &
    \editimg{#1/sample\_7\_x\_hat.png} &
    \editimg{#1/sample\_8\_x\_hat.png} &
    \editimg{#1/sample\_9\_x\_hat.png} &
    \editimg{#1/sample\_10\_x\_hat.png} &
    \editimg{#1/sample\_11\_x\_hat.png}%
}

\newcommand{\weighteditrow}[1]{%
    \editimg{#1/sample\_0\_x\_hat.png} &
    \editimg{#1/sample\_1\_x\_hat.png} &
    \editimg{#1/sample\_2\_x\_hat.png} &
    \editimg{#1/sample\_3\_x\_hat.png} &
    \editimg{#1/sample\_4\_x\_hat.png} &
    \editimg{#1/sample\_5\_x\_hat.png} &
    \editimg{#1/sample\_6\_x\_hat.png} &
    \editimg{#1/sample\_7\_x\_hat.png} &
    \editimg{#1/sample\_8\_x\_hat.png} &
    \editimg{#1/sample\_9\_x\_hat.png}%
}

\begin{tabular}{@{}cccccccccc@{}}

    \multicolumn{10}{c}{\scriptsize\textbf{Latent-Space Edit}} \\
    \scriptsize $\gamma=0$ &
    \scriptsize $\gamma=0.44$ &
    \scriptsize $\gamma=0.46$ &
    \scriptsize $\gamma=0.48$ &
    \scriptsize $\gamma=0.50$ &
    \scriptsize $\gamma=0.52$ &
    \scriptsize $\gamma=0.54$ &
    \scriptsize $\gamma=0.56$ &
    \scriptsize $\gamma=0.58$ &
    \scriptsize $\gamma=0.60$
    \\[2pt]
    \latenteditrow{figures/06\_edit/edit\_Bald\_sample1\_bald}
    \\[3pt]

    \multicolumn{10}{c}{%
        \scriptsize\textbf{Weight-Space Edit}} \\
    \scriptsize $\gamma=0$ &
    \scriptsize $\gamma=4$ &
    \scriptsize $\gamma=5$ &
    \scriptsize $\gamma=5.33$ &
    \scriptsize $\gamma=5.66$ &
    \scriptsize $\gamma=6$ &
    \scriptsize $\gamma=7$ &
    \scriptsize $\gamma=8$ &
    \scriptsize $\gamma=9$ &
    \scriptsize $\gamma=10$
    \\[2pt]
    \weighteditrow{figures/06\_edit/wedit\_Bald\_sample1}

\end{tabular}

\caption{
Editing along the fitted ridge direction produces the same full-hair-to-bald transition whether applied to the latent (top) or directly to the raw adapter weights (bottom). The two rows use different $\gamma$ ranges since the direction's scale differs between the two spaces.
}
\label{fig:editing-comparison}
\end{figure}

\paragraph{Results.}
\begin{wrapfigure}[23]{r}{0.55\columnwidth}
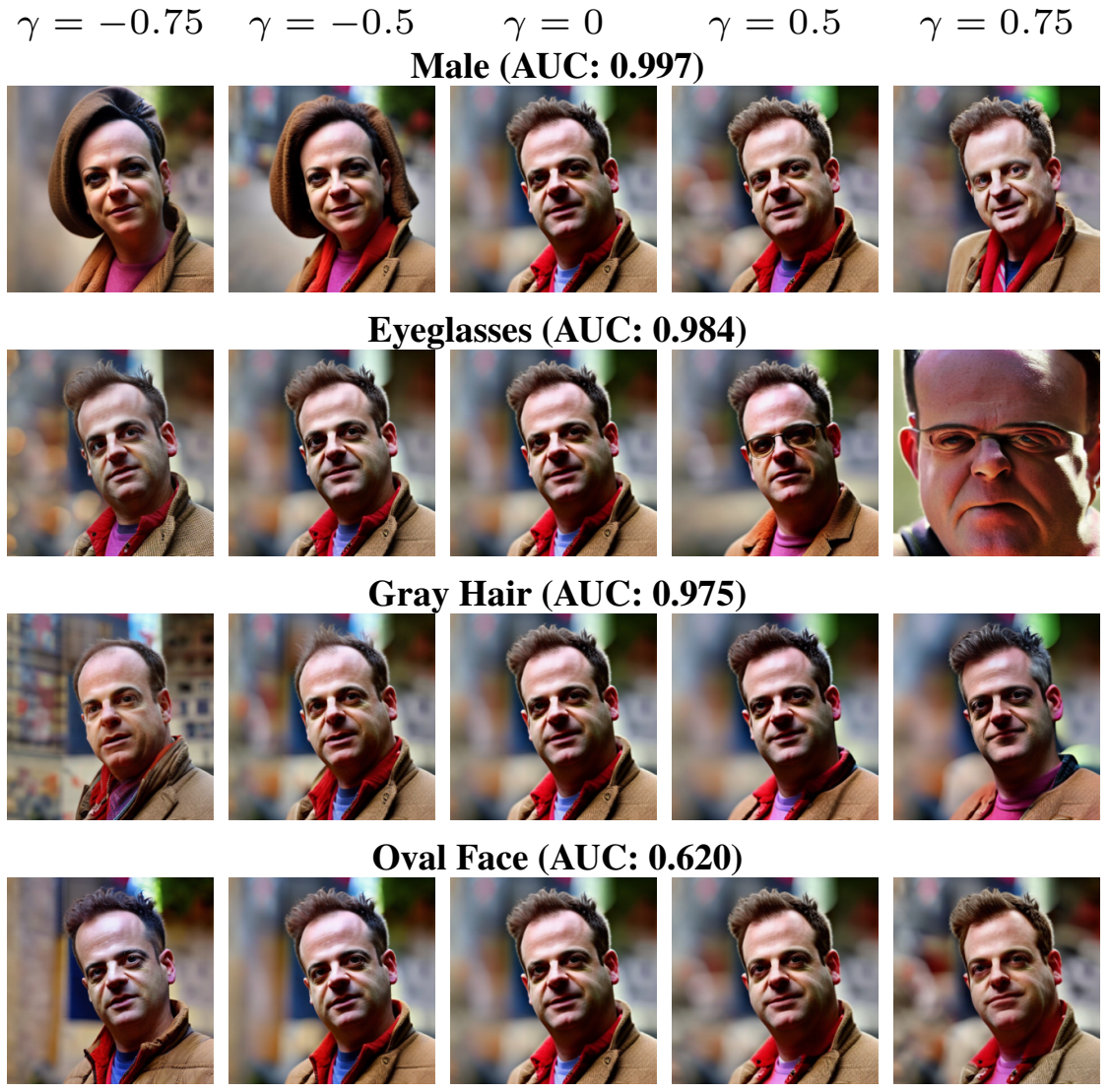

\centering
\vspace{-15pt}

\setlength{\tabcolsep}{1.5pt}
\renewcommand{\arraystretch}{0.5}

\newcommand{\editimg}[1]{%
    \includegraphics[width=0.18\linewidth,height=0.18\linewidth]{#1}%
}

\newcommand{\editrow}[1]{%
    \editimg{#1/sample\_0\_x\_hat.png} &
    \editimg{#1/sample\_1\_x\_hat.png} &
    \editimg{#1/sample\_3\_x\_hat.png} &
    \editimg{#1/sample\_5\_x\_hat.png} &
    \editimg{#1/sample\_6\_x\_hat.png}%
}

\begin{tabular}{@{}*{7}{c}@{}}
    \scriptsize $\gamma=-0.75$ &
    \scriptsize $\gamma=-0.5$ &
    \scriptsize $\gamma=0$ &
    \scriptsize $\gamma=0.5$ &
    \scriptsize $\gamma=0.75$ \\[2pt]

    \multicolumn{5}{c}{\scriptsize\textbf{Male (AUC: 0.997)}} \\
    \editrow{figures/06\_edit/edit\_Male\_sample5} \\[3pt]

    \multicolumn{5}{c}{\scriptsize\textbf{Eyeglasses (AUC: 0.984)}} \\
    \editrow{figures/06\_edit/edit\_Eyeglasses\_sample5} \\[3pt]

    \multicolumn{5}{c}{\scriptsize\textbf{Gray Hair (AUC: 0.975)}} \\
    \editrow{figures/06\_edit/edit\_Gray\_Hair\_sample5} \\[3pt]

    \multicolumn{5}{c}{\scriptsize\textbf{Oval Face (AUC: 0.620)}} \\
    \editrow{figures/06\_edit/edit\_Oval\_Face\_sample5}
\end{tabular}
\caption{
Edit quality declines with the direction's held-out AUC, from clean edits for \emph{Male} and \emph{Eyeglasses} to a barely visible change for \emph{Oval Face}.
}
\label{fig:steering-comparison}
\vspace{-6pt}
\end{wrapfigure}
On \emph{Bald}, one of the rarest attributes in the dataset ($2.5\%$), the flattened edit still produces a clean, continuous sweep (Figure~\ref{fig:editing-comparison}). At $\gamma = 0$ the reconstruction renders the unedited identity, and as $\gamma$ increases the hairline recedes and then disappears into a fully bald head, with intermediate strengths producing intermediate hairlines and the rest of the identity, skin tone, expression, and framing, preserved throughout. The identical ridge direction fitted directly on the raw adapter weights and applied without the autoencoder produces the same qualitative result over a comparable sweep (Figure~\ref{fig:editing-comparison}). This matches what w2w~\citep{dravid2024interpreting} already showed for a linear direction fitted post hoc on raw LoRA weights, here with ridge in place of a truncated-PCA pseudo-inverse. Applying the same construction to four further attributes on held-out identities (Figure~\ref{fig:steering-comparison}, more examples in Figure~\ref{fig:steering-comparison-extended} in Appendix~\ref{app:editing}) shows edit quality tracking the AUC score fairly well. \emph{Male} and \emph{Eyeglasses} score highest and edit the most cleanly, the former moving continuously between a feminine and a masculine rendering with hair length, jawline, and facial hair changing together, the latter adding glasses as a discrete object. \emph{Gray Hair} follows close behind with a milder but still clearly visible lightening of the hair, while \emph{Oval Face}, well below the others, produces an edit that barely registers.

\section{The Latent Space Improves Robustness}
\label{sec:ex-part2}

This section tests the latent under two kinds of displacement: a single perturbation of varying size, compared directly against the same perturbation applied to the raw weights, and repeated encode--decode passes on the same adapter, which is what happens in practice whenever a sampled or edited latent is fed back through the autoencoder.

\subsection{Perturbation}

\emph{How far can an adapter be perturbed before the identity breaks down, and does it fail gracefully or collapse abruptly?}

\paragraph{Setup.} We test how far an adapter can be moved, in either space, before the generated identity breaks down, and whether it degrades smoothly or collapses abruptly. Each perturbation is scaled by the standard deviation of the space it acts in, so a noise scale $\alpha$ denotes the same self-relative displacement, a fraction $\alpha$ of that space's own spread. In the latent we form $z' = z + \alpha\,\hat\sigma_z\,\varepsilon$ for a real identity's latent $z = E(\theta)$, with $\varepsilon \sim \mathcal{N}(0, I)$ and $\hat\sigma_z$ the global standard deviation of the latent. In weight space we form the analogous $\theta' = \theta + \alpha\,\hat\sigma_\theta\,\varepsilon$, with $\hat\sigma_\theta$ the global standard deviation of the adapter population. We sweep $\alpha$ over the same range in both spaces.

\begin{figure}[t]
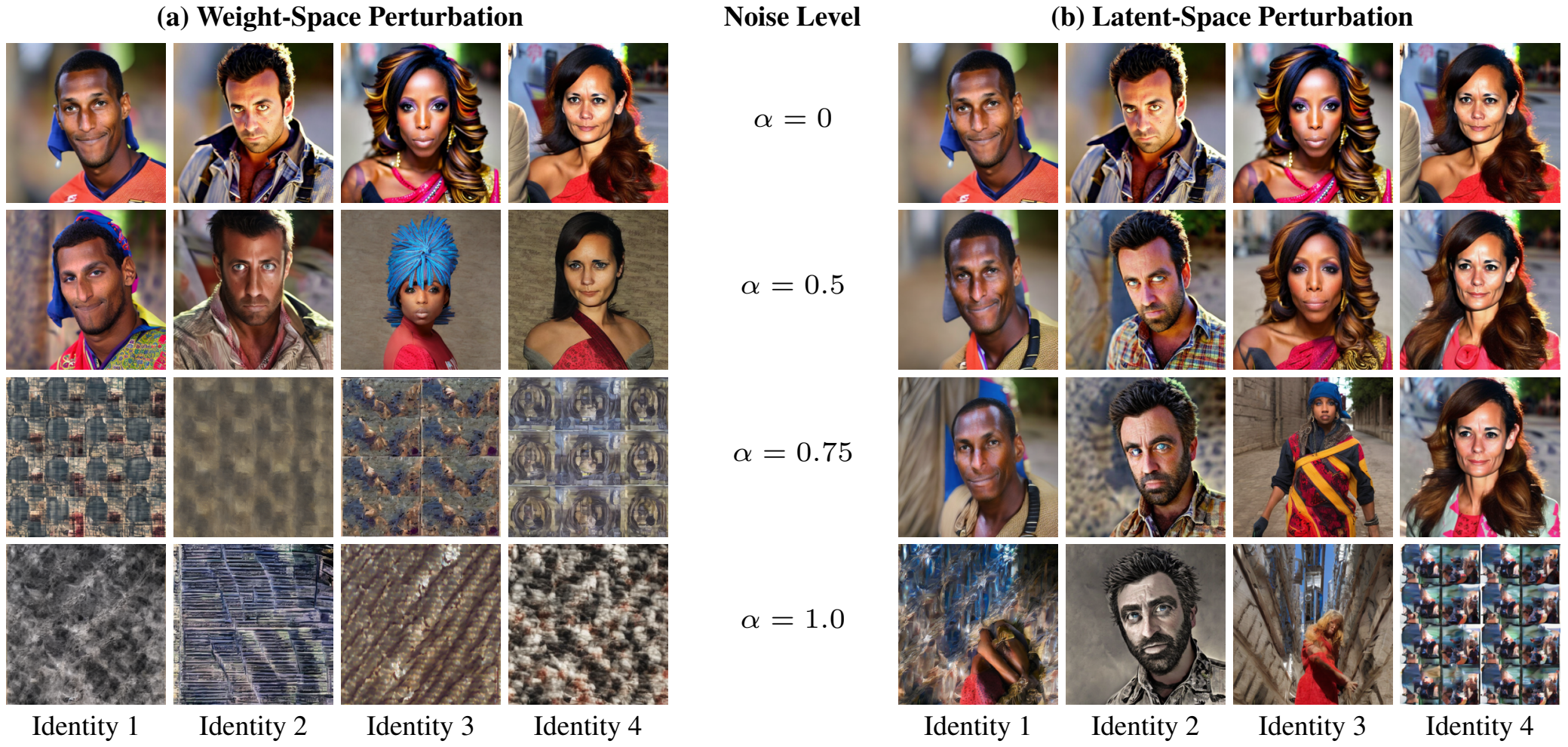

\centering

\newlength{\gridimgwidth}
\setlength{\gridimgwidth}{0.1\linewidth}

\newcommand{\cellimg}[1]{%
    \includegraphics[width=\gridimgwidth]{#1}%
}

\newsavebox{\alpharefbox}

\newcommand{\alphacell}[1]{%
    \sbox{\alpharefbox}{%
        \cellimg{figures/03\_weight\_noise/noisy\_000/sample\_5738\_x.png}%
    }%
    \parbox[b][\ht\alpharefbox][c]{\linewidth}{%
        \centering\scriptsize $#1$%
    }%
}

\begin{minipage}[t]{0.44\linewidth}
\centering
\scriptsize\textbf{(a) Weight-Space Perturbation}\par\vspace{2pt}

{\setlength{\tabcolsep}{1pt}
\renewcommand{\arraystretch}{0.4}
\begin{tabular}{@{}cccc@{}}
    \cellimg{figures/03\_weight\_noise/noisy\_000/sample\_5738\_x.png} &
    \cellimg{figures/03\_weight\_noise/noisy\_000/sample\_5740\_x.png} &
    \cellimg{figures/03\_weight\_noise/noisy\_000/sample\_5742\_x.png} &
    \cellimg{figures/03\_weight\_noise/noisy\_000/sample\_5744\_x.png} \\[1pt]

    \cellimg{figures/03\_weight\_noise/noisy\_050/sample\_5738\_x.png} &
    \cellimg{figures/03\_weight\_noise/noisy\_050/sample\_5740\_x.png} &
    \cellimg{figures/03\_weight\_noise/noisy\_050/sample\_5742\_x.png} &
    \cellimg{figures/03\_weight\_noise/noisy\_050/sample\_5744\_x.png} \\[1pt]

    \cellimg{figures/03\_weight\_noise/noisy\_075/sample\_5738\_x.png} &
    \cellimg{figures/03\_weight\_noise/noisy\_075/sample\_5740\_x.png} &
    \cellimg{figures/03\_weight\_noise/noisy\_075/sample\_5742\_x.png} &
    \cellimg{figures/03\_weight\_noise/noisy\_075/sample\_5744\_x.png} \\[1pt]

    \cellimg{figures/03\_weight\_noise/noisy\_100/sample\_5738\_x.png} &
    \cellimg{figures/03\_weight\_noise/noisy\_100/sample\_5740\_x.png} &
    \cellimg{figures/03\_weight\_noise/noisy\_100/sample\_5742\_x.png} &
    \cellimg{figures/03\_weight\_noise/noisy\_100/sample\_5744\_x.png} \\[2pt]

    \scriptsize Identity 1 & \scriptsize Identity 2 &
    \scriptsize Identity 3 & \scriptsize Identity 4 \\
\end{tabular}
}
\end{minipage}
\hfill
\begin{minipage}[t]{0.10\linewidth}
\centering
\scriptsize\textbf{Noise Level}\par\vspace{2pt}

{\setlength{\tabcolsep}{0pt}
\renewcommand{\arraystretch}{0.4}
\begin{tabular}{c}
    \alphacell{\alpha=0} \\[1pt]
    \alphacell{\alpha=0.5} \\[1pt]
    \alphacell{\alpha=0.75} \\[1pt]
    \alphacell{\alpha=1.0}
\end{tabular}
}
\end{minipage}
\begin{minipage}[t]{0.44\linewidth}
\centering
\scriptsize\textbf{(b) Latent-Space Perturbation}\par\vspace{2pt}

{\setlength{\tabcolsep}{1pt}
\renewcommand{\arraystretch}{0.4}
\begin{tabular}{@{}cccc@{}}
    \cellimg{figures/03\_latent\_noise/0/sample\_4\_x.png} &
    \cellimg{figures/03\_latent\_noise/0/sample\_6\_x.png} &
    \cellimg{figures/03\_latent\_noise/0/sample\_8\_x.png} &
    \cellimg{figures/03\_latent\_noise/0/sample\_10\_x.png} \\[1pt]

    \cellimg{figures/03\_latent\_noise/05/sample\_4\_x\_hat.png} &
    \cellimg{figures/03\_latent\_noise/05/sample\_6\_x\_hat.png} &
    \cellimg{figures/03\_latent\_noise/05/sample\_8\_x\_hat.png} &
    \cellimg{figures/03\_latent\_noise/05/sample\_10\_x\_hat.png} \\[1pt]

    \cellimg{figures/03\_latent\_noise/075/sample\_4\_x\_hat.png} &
    \cellimg{figures/03\_latent\_noise/075/sample\_6\_x\_hat.png} &
    \cellimg{figures/03\_latent\_noise/075/sample\_8\_x\_hat.png} &
    \cellimg{figures/03\_latent\_noise/075/sample\_10\_x\_hat.png} \\[1pt]

    \cellimg{figures/03\_latent\_noise/1/sample\_4\_x\_hat.png} &
    \cellimg{figures/03\_latent\_noise/1/sample\_6\_x\_hat.png} &
    \cellimg{figures/03\_latent\_noise/1/sample\_8\_x\_hat.png} &
    \cellimg{figures/03\_latent\_noise/1/sample\_10\_x\_hat.png} \\[2pt]

    \scriptsize Identity 1 & \scriptsize Identity 2 &
    \scriptsize Identity 3 & \scriptsize Identity 4 \\
\end{tabular}
}
\end{minipage}

\caption{
Perturbations in (a) weight-space and (b) latent-space across increasing noise levels $\alpha$. Each column depicts one identity under the same perturbation level.
}
\label{fig:perturbation-comparison}
\end{figure}

\paragraph{Results.} Weight-space perturbation degrades the identity abruptly (Figure~\ref{fig:perturbation-comparison}, left). At $\alpha=0.5$ identities remain broadly recognizable, though some detail is already lost. By $\alpha=0.75$ most have collapsed into texture-like, off-distribution images that are no longer faces. At $\alpha=1.0$ every identity has dissolved into noise. The transition is sharp, with little intermediate regime in which the adapter is perturbed yet still valid. Perturbing the latent by the same fractional amount degrades far more gracefully (Figure~\ref{fig:perturbation-comparison}, right). At $\alpha=0.5$ identities are essentially preserved. At $\alpha=0.75$, where the weight-space adapters have already collapsed, latent-perturbed identities remain clearly recognizable despite larger variation. Only at $\alpha=1.0$ does the identity begin to break down, and even then it tends to degrade into a different or distorted face rather than into the texture-like noise seen in weight space. The raw weight space supports valid adapters only within a narrow band around each trained adapter. The neighborhood of a real point in the latent space, by contrast, is populated by valid, related identities.

\newpage
\subsection{Reconstruction Drift}

\emph{Does the small error of a single encode--decode pass compound under repetition, or does it settle to a fixed point?}

\paragraph{Setup.} We repeatedly apply the encoder and decoder to the same adapter. Starting from a real adapter $\theta^{(0)} = \theta$, we form $\theta^{(k)} = D(E(\theta^{(k-1)}))$ for $k = 1, \dots, 10$, applying each $\theta^{(k)}$ to the frozen base model and rendering it under the fixed prompt and seed used throughout. We track $R^2$ between $\theta^{(k)}$ and the original $\theta^{(0)}$ and the LPIPS distance between their renders. A per-step change that decays toward zero would indicate a fixed point, while a change that stays roughly constant would indicate unbounded accumulation.

\begin{figure*}[t]
\centering

\begin{minipage}[t]{0.32\textwidth}
\vspace{0pt}
\centering

\scriptsize\textbf{(a) Weight-Space $R^2$ vs.\ Original}\par\vspace{2pt}
\begin{tikzpicture}
\begin{axis}[
    cyclepanel,
    width=\linewidth,
    height=3.5cm,
    ylabel={$R^2$},
    ymin=0.933, ymax=1.006,
    ytick={0.94,0.96,0.98,1.00},
    yticklabel style={
        /pgf/number format/fixed,
        /pgf/number format/precision=2
    },
    legend style={
        draw=none, fill=none, font=\footnotesize,
        at={(0.97,0.97)}, anchor=north east
    },
]
    \addplot[draw=none, forget plot, name path=rsqhi] coordinates {
        (0,1.00000000) (1,0.99370365) (2,0.99187753) (3,0.98931568)
        (4,0.98614353) (5,0.98244009) (6,0.97820855) (7,0.97332507)
        (8,0.96742774) (9,0.96028630) (10,0.95204827)
    };
    \addplot[draw=none, forget plot, name path=rsqlo] coordinates {
        (0,1.00000000) (1,0.99299093) (2,0.99089623) (3,0.98786294)
        (4,0.98398569) (5,0.97920247) (6,0.97318163) (7,0.96552235)
        (8,0.95652354) (9,0.94689812) (10,0.93704195)
    };
    \addplot[bandstyle, forget plot] fill between[of=rsqhi and rsqlo];
    \addplot[meanstyle] coordinates {
        (0,1.00000000) (1,0.99334729) (2,0.99138688) (3,0.98858931)
        (4,0.98506461) (5,0.98082128) (6,0.97569509) (7,0.96942371)
        (8,0.96197564) (9,0.95359221) (10,0.94454511)
    };
\end{axis}
\end{tikzpicture}

\vspace{4pt}

\scriptsize\textbf{(b) LPIPS vs.\ Original Image}\par\vspace{2pt}
\begin{tikzpicture}
\begin{axis}[
    cyclepanel,
    width=\linewidth,
    height=3.5cm,
    ylabel={LPIPS},
    ymin=-0.03, ymax=0.62,
    ytick={0,0.2,0.4,0.6},
    yticklabel style={
        /pgf/number format/fixed,
        /pgf/number format/precision=1
    },
]
    \addplot[draw=none, forget plot, name path=lphi] coordinates {
        (0,0.00000000) (1,0.38737507) (2,0.40099934) (3,0.42954417)
        (4,0.44581487) (5,0.45379191) (6,0.46169420) (7,0.46828943)
        (8,0.47234735) (9,0.47778315) (10,0.48670798)
    };
    \addplot[draw=none, forget plot, name path=lplo] coordinates {
        (0,0.00000000) (1,0.07431425) (2,0.09017582) (3,0.10490257)
        (4,0.12856839) (5,0.15065257) (6,0.17074646) (7,0.19029247)
        (8,0.20798169) (9,0.21796531) (10,0.23217922)
    };
    \addplot[bandstyle, forget plot] fill between[of=lphi and lplo];
    \addplot[meanstyle] coordinates {
        (0,0.00000000) (1,0.23084466) (2,0.24558758) (3,0.26722337)
        (4,0.28719163) (5,0.30222224) (6,0.31622033) (7,0.32929095)
        (8,0.34016452) (9,0.34787423) (10,0.35944360)
    };
\end{axis}
\end{tikzpicture}
\end{minipage}
\hfill
\begin{minipage}[t]{0.66\textwidth}
\vspace{0pt}
\centering
\scriptsize\textbf{(c) Selected Encode-Decode Cycles}\par\vspace{2pt}

\newcommand{\cycleimg}[1]{%
    \includegraphics[width=0.155\linewidth,height=0.155\linewidth]{#1}%
}
{\setlength{\tabcolsep}{1.5pt}
\renewcommand{\arraystretch}{0.5}
\begin{tabular}{@{}*{6}{c}@{}}
    \scriptsize $k=0$ &
    \scriptsize $k=1$ &
    \scriptsize $k=3$ &
    \scriptsize $k=5$ &
    \scriptsize $k=7$ &
    \scriptsize $k=10$ \\[2pt]

    \cycleimg{figures/07\_cycle/round\_00/sample\_0\_x\_hat.png} &
    \cycleimg{figures/07\_cycle/round\_01/sample\_0\_x\_hat.png} &
    \cycleimg{figures/07\_cycle/round\_03/sample\_0\_x\_hat.png} &
    \cycleimg{figures/07\_cycle/round\_05/sample\_0\_x\_hat.png} &
    \cycleimg{figures/07\_cycle/round\_07/sample\_0\_x\_hat.png} &
    \cycleimg{figures/07\_cycle/round\_10/sample\_0\_x\_hat.png} \\[1pt]

    \cycleimg{figures/07\_cycle/round\_00/sample\_1\_x\_hat.png} &
    \cycleimg{figures/07\_cycle/round\_01/sample\_1\_x\_hat.png} &
    \cycleimg{figures/07\_cycle/round\_03/sample\_1\_x\_hat.png} &
    \cycleimg{figures/07\_cycle/round\_05/sample\_1\_x\_hat.png} &
    \cycleimg{figures/07\_cycle/round\_07/sample\_1\_x\_hat.png} &
    \cycleimg{figures/07\_cycle/round\_10/sample\_1\_x\_hat.png} \\[1pt]

    \cycleimg{figures/07\_cycle/round\_00/sample\_2\_x\_hat.png} &
    \cycleimg{figures/07\_cycle/round\_01/sample\_2\_x\_hat.png} &
    \cycleimg{figures/07\_cycle/round\_03/sample\_2\_x\_hat.png} &
    \cycleimg{figures/07\_cycle/round\_05/sample\_2\_x\_hat.png} &
    \cycleimg{figures/07\_cycle/round\_07/sample\_2\_x\_hat.png} &
    \cycleimg{figures/07\_cycle/round\_10/sample\_2\_x\_hat.png} \\[1pt]

    \cycleimg{figures/07\_cycle/round\_00/sample\_3\_x\_hat.png} &
    \cycleimg{figures/07\_cycle/round\_01/sample\_3\_x\_hat.png} &
    \cycleimg{figures/07\_cycle/round\_03/sample\_3\_x\_hat.png} &
    \cycleimg{figures/07\_cycle/round\_05/sample\_3\_x\_hat.png} &
    \cycleimg{figures/07\_cycle/round\_07/sample\_3\_x\_hat.png} &
    \cycleimg{figures/07\_cycle/round\_10/sample\_3\_x\_hat.png} \\
\end{tabular}}
\end{minipage}

\caption{Stability under repeated encode--decode cycles. (a) Weight-space similarity to the original adapter decreases gradually. (b) Rendered images increasingly differ from the original, measured by LPIPS. Shaded regions denote $\pm1\sigma$. (c) Representative renders after $k=0,1,3,5,7,10$ cycles.}
\label{fig:drift}
\end{figure*}
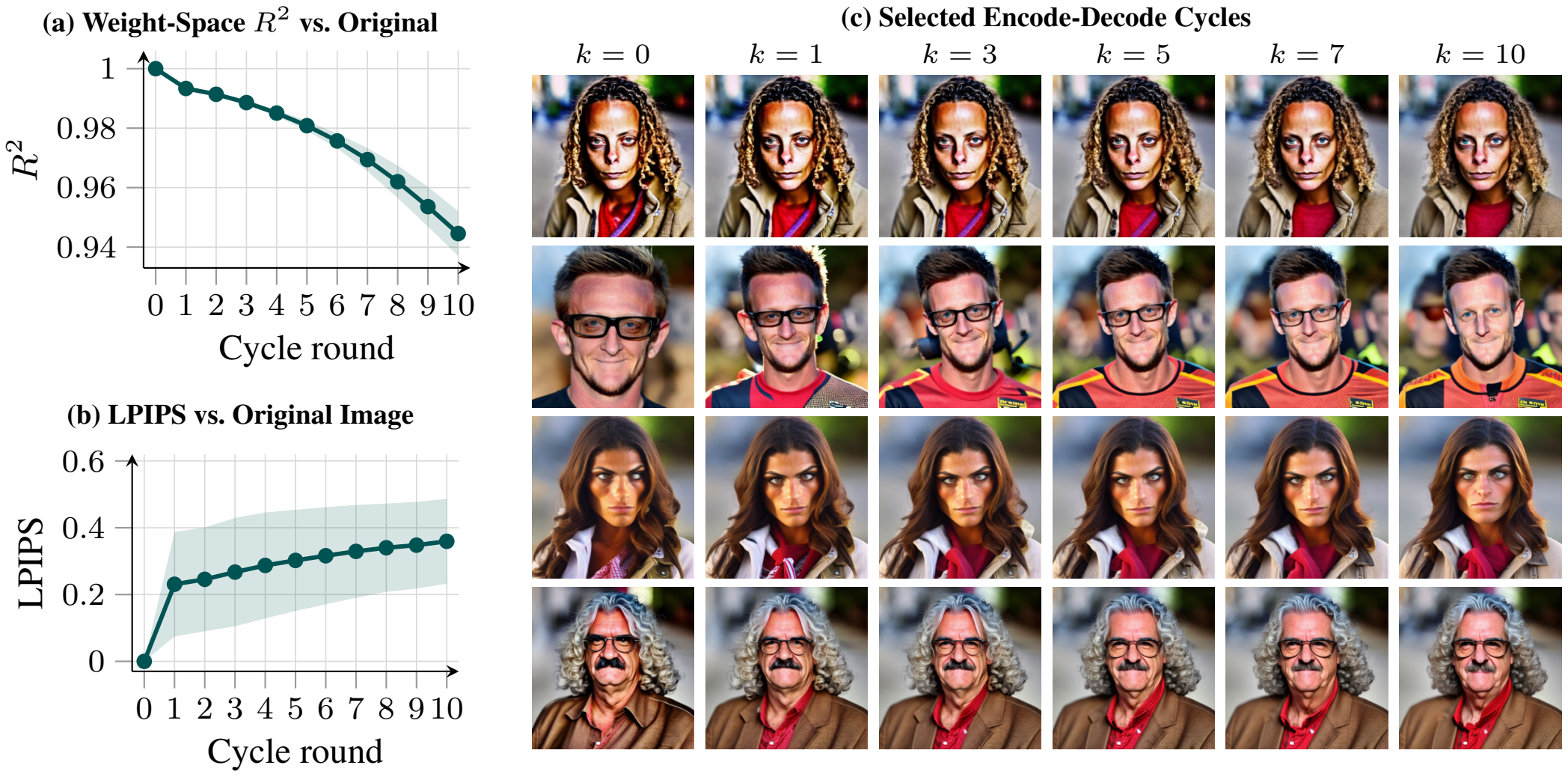

\paragraph{Results.} Identities survive all ten cycles (Figure~\ref{fig:drift}). Facial structure, hair, coloring, expression, and framing remain intact, and almost all visible change happens at the first cycle alone. The two quantitative measures capture different sides of this. LPIPS saturates, rising to $0.23$ after one cycle and reaching only $0.36$ after ten. $R^2$ declines steadily from $1$ to about $0.945$, and the per-cycle loss grows rather than shrinks, from $0.004$ over the first five cycles to $0.007$ over the second five. Ten cycles are safe in practice, since the renders stay recognizable and the perceptual cost of each additional cycle is small, but that safety reflects the renderer's tolerance rather than the autoencoder converging. The weight-space deviation gives no sign of stopping, so extrapolating to many more cycles is not supported by this result.

\section{The Latent Space Generalizes Across Adapter Ranks}
\label{sec:ex-part3}

\textit{Can a single autoencoder absorb adapters of varying ranks, treating adapter weights as a general modality rather than one tied to a fixed architecture?}

\paragraph{Setup.} Doubling the LoRA rank doubles adapter dimensionality from $d = 99{,}648$ to $d = 199{,}296$, so a rank-2 adapter does not fit the token sequence a rank-1 model expects. Our dense tokenizer assigns each token a fixed range of weight indices. At rank 2, the same token index falls in a different part of the network than it does at rank 1. The positional encoding the encoder relies on would therefore become misleading. We instead use a boundary-respecting sparse tokenizer at width $640$, sized to a matrix's rank-2 parameter count. A rank-1 matrix then occupies only the first half of its token budget, with the remainder marked as padding. Token position refers to the same matrix at both ranks. We train one autoencoder on a pool containing both rank-1 and rank-2 adapters and compare it against the rank-1-only model used throughout the rest of our experiments, reporting reconstruction quality as LPIPS on held-out adapters of each rank.

\begin{figure}[t]
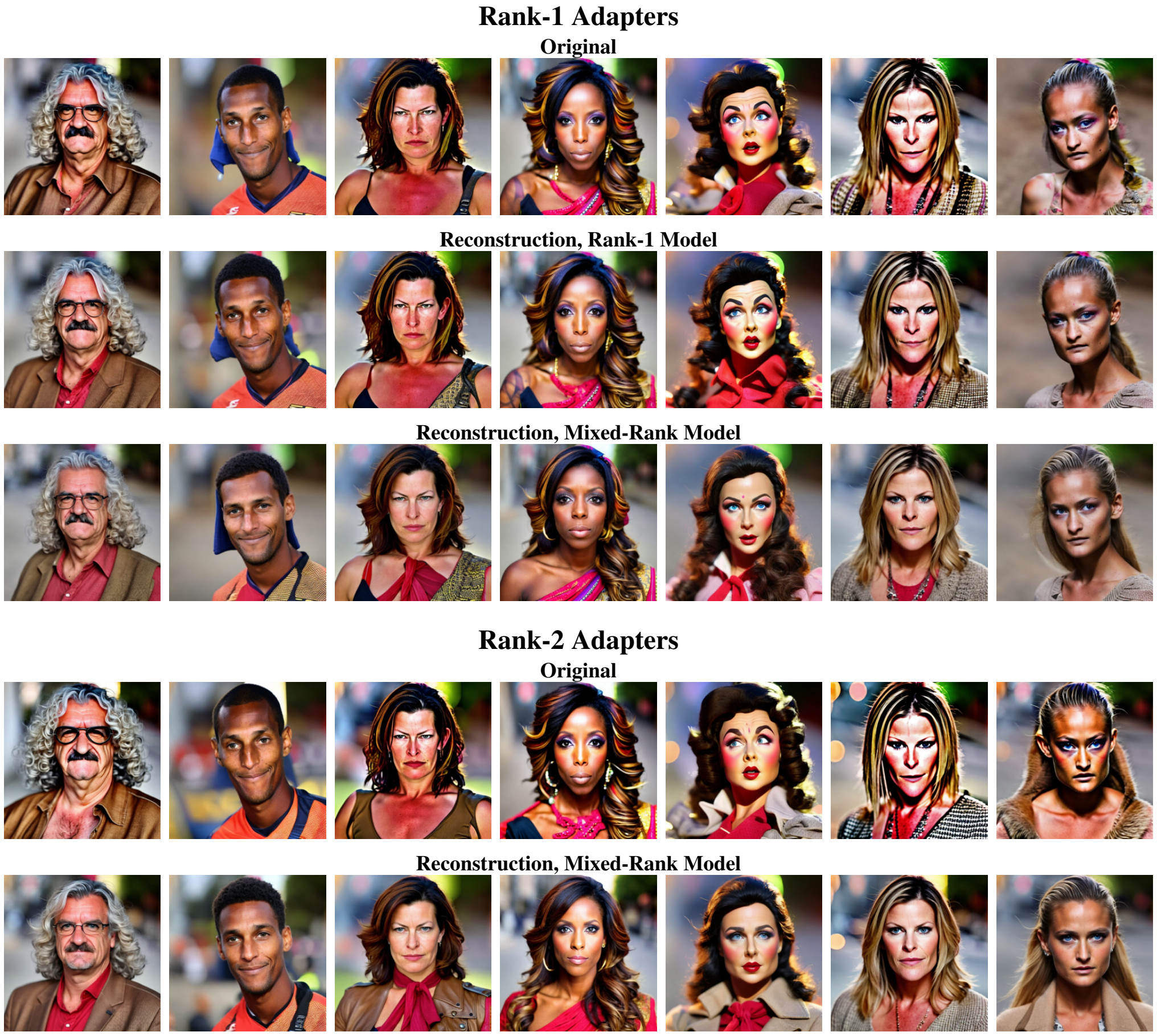

\centering

\setlength{\tabcolsep}{1.5pt}
\renewcommand{\arraystretch}{0.5}

\newcommand{\reconimg}[1]{%
    \includegraphics[width=0.13\linewidth,height=0.13\linewidth]{#1}%
}

\newcommand{\rankonerow}[2]{%
    \reconimg{#1/sample\_3\_#2.png} &
    \reconimg{#1/sample\_4\_#2.png} &
    \reconimg{#1/sample\_7\_#2.png} &
    \reconimg{#1/sample\_8\_#2.png} &
    \reconimg{#1/sample\_13\_#2.png} &
    \reconimg{#1/sample\_15\_#2.png} &
    \reconimg{#1/sample\_16\_#2.png}%
}

\newcommand{\ranktworow}[2]{%
    \reconimg{#1/sample\_9392\_#2.png} &
    \reconimg{#1/sample\_9393\_#2.png} &
    \reconimg{#1/sample\_9394\_#2.png} &
    \reconimg{#1/sample\_9395\_#2.png} &
    \reconimg{#1/sample\_9398\_#2.png} &
    \reconimg{#1/sample\_9399\_#2.png} &
    \reconimg{#1/sample\_9400\_#2.png}%
}

\begin{tabular}{@{}ccccccc@{}}

    \multicolumn{7}{c}{\small\textbf{Rank-1 Adapters}} \\[3pt]

    \multicolumn{7}{c}{\scriptsize\textbf{Original}} \\
    \rankonerow
        {figures/09\_mixed\_rank/dense\_l256\_epoch800\_val}
        {x}
    \\[4pt]

    \multicolumn{7}{c}{\scriptsize\textbf{Reconstruction, Rank-1 Model}} \\
    \rankonerow
        {figures/09\_mixed\_rank/dense\_l256\_epoch800\_val}
        {x\_hat}
    \\[4pt]

    \multicolumn{7}{c}{\scriptsize\textbf{Reconstruction, Mixed-Rank Model}} \\
    \rankonerow
        {figures/09\_mixed\_rank/combined\_r1r2\_sparse\_l256\_d640\_h10\_l6\_r1}
        {x\_hat}
    \\[8pt]

    \multicolumn{7}{c}{\small\textbf{Rank-2 Adapters}} \\[3pt]

    \multicolumn{7}{c}{\scriptsize\textbf{Original}} \\
    \ranktworow
        {figures/09\_mixed\_rank/combined\_r1r2\_sparse\_l256\_d640\_h10\_l6\_r2}
        {x}
    \\[4pt]

    \multicolumn{7}{c}{\scriptsize\textbf{Reconstruction, Mixed-Rank Model}} \\
    \ranktworow
        {figures/09\_mixed\_rank/combined\_r1r2\_sparse\_l256\_d640\_h10\_l6\_r2}
        {x\_hat}

\end{tabular}

\caption{
A single mixed-rank autoencoder reconstructs recognizable identities at both ranks. Rows 1–3 compare rank-1 adapters reconstructed by the rank-specific model against the mixed-rank model, and rows 4–5 show rank-2 adapters reconstructed by the mixed-rank model.
}
\label{fig:mixedrank-comparison}
\end{figure}

\paragraph{Results.} A single autoencoder, with no increase in latent size or model capacity, reconstructs recognizable identities at both ranks (Figure~\ref{fig:mixedrank-comparison}). The rank-2 half of every token is therefore genuinely learned rather than ignored, and a shared latent over two adapter ranks is achievable within the same backbone used throughout the rest of our experiments. This capacity is now split between two ranks rather than dedicated to one, and the cost shows up in fidelity. On rank-1 adapters the mixed model reaches an average LPIPS of $0.29$ against $0.22$ for the rank-1-only model, about $32\%$ worse, and it reaches $0.35$ on rank-2 adapters. Section~\ref{sec:method-eval} shows that weight-space $R^2$ and perceptual quality track each other closely, so this gap reflects a loss of reconstruction and places the mixed model in the range we observed for rank-1 adapters before convergence.

\section{Limitations}
\label{sec:limitations}

\paragraph{Splitting.} Train, validation, and test splits are drawn per adapter rather than per identity, so near-duplicate fits of the same person can appear across different splits. This does not affect the qualitative renders that carry most of our claims, but it makes held-out numbers such as the AUC optimistic relative to a strict identity-level split, since a direction can be scored against adapters that are near-duplicates of ones it effectively saw in training. Identity-level splitting is a straightforward fix left to future work.

\paragraph{Reconstruction floor.} The best rank-1 model reconstructs adapters to an average LPIPS of about $0.22$, and the first encode-decode cycle in our drift experiment costs $0.23$, almost the same number, and larger than the drift accumulated over the following nine cycles combined. Every latent operation in this paper is therefore observed through a reconstruction already perceptibly different from its source, which is why the $\gamma = 0$ control in each edit sweep is $D(E(\theta))$ rather than $\theta$ itself. w2w, which samples and edits directly in weight space, has no equivalent floor.

\paragraph{Scope.} Several limits on scope apply throughout. Everything rests on rank-1 and rank-2 updates to the query and value projections of a single frozen backbone, trained by the same DreamBooth-style procedure, so no claim about adapter weights as a general modality follows from this paper alone. Attribute editing also does not strictly require the learned latent, since the same ridge direction fitted directly on raw adapters produces a comparable edit, reproducing w2w in a different basis. The attribute fits themselves are underdetermined. Training uses about $5{,}734$ adapters against a flattened latent of $79{,}872$ dimensions, so the ridge fits interpolate the labels even at $\lambda = 1$, and the fitted direction itself is not uniquely identified by the data.

\paragraph{Out-of-distribution generalization.} We test generalization across LoRA rank but not across subject domain. The autoencoder is trained exclusively on adapters encoding human identities, and whether its latent organizes adapters more broadly, for instance the dog-breed adapters also released by w2w, or has specialized to faces specifically, is a question this paper leaves open for future work.

\section{Discussion}
\label{sec:discussion}

A structured latent over adapters mainly changes cost rather than capability. A new identity becomes a decoder pass instead of a training run, and a new attribute becomes one closed-form solve against an existing Gram matrix instead of a dedicated LoRA. More broadly, our experiments show that the different operations possible in the LoRA weight space are preserved in the learned latent, with two distinct advantages beyond preservation. The first advantage appears under displacement. Perturbations of comparable magnitude in weight space cause considerable degradation of the identity, while the learned representations show greater robustness and still decode into recognizable identities. This behavior is particularly relevant for downstream operations such as optimization and generation, where greater robustness to perturbations is necessary. The second primary advantage is the possibility to learn representations of different rank adapters. Operations in the raw adapter weight space require a shared dimensionality, whereas our tokenized approach represents different rank adapters within a common latent space.

Taken together, these results show that a learned LoRA latent space provides a more robust and extensible representation for performing multiple different downstream operations. Whether this advantage hold across different domains and tasks remains a key question for moving towards adapter weights as a general data modality.

\section{Broader Impact}
\label{sec:impact}
The identities studied here are synthetic, but nothing in the method depends on that, and composing this latent with an inversion step like w2w's, turning a single photograph into a point in weight space, would in principle extend it to real people, letting a specific individual's apparent age or gender presentation be moved continuously while their identity is preserved. This kind of consistency is exactly what makes fake personas cheap at scale. A second concern sits in the attribute directions themselves. Several CelebA attributes are annotator judgments rather than observable properties, and fitting a direction to one turns a cultural judgment into a coordinate that can be adjusted continuously on a real person's face. This paper uses only synthetic identities and does not implement inversion, which keeps it clear of the most direct misuse path, though that is a property of the experiments rather than a guarantee of the method. We take the position that adapters encoding real people should require consent at the level of the identity, obtained beforehand.

\newpage
\bibliographystyle{plainnat}
\bibliography{references}


\newpage
\appendix

\section{Implementation Details}

\label{app:implementation}
 
\paragraph{Adapter instantiation.} LoRA is inserted into the query and value projections of both self- and cross-attention in the U-Net, while the key projections and all other weights stay frozen. Each pair on an image-space projection contributes $2C$ parameters at channel width $C$, and three of the four adapted projections are of this kind. Cross-attention's value takes the $768$-dimensional CLIP text embedding as input, adding a further $768$ parameters. Summed over one block this gives $7C + 768$ parameters, and over the $16$ attention blocks spanning $C \in \{320, 640, 1280\}$ (Table~\ref{tab:lora-param-count}), a total adapter dimensionality of $d = 99{,}648$ at rank $1$. Doubling the rank doubles the per-block count to $14C + 1{,}536$ and the total to $d = 199{,}296$. The rank-2 adapters used in Section~\ref{sec:ex-part3} are generated with the original training pipeline released by \citet{dravid2024interpreting}, changing only the LoRA rank from 1 to 2. The base model, DreamBooth objective, and prompt template are otherwise identical to the released rank-1 recipe. Our released repository operates on the resulting adapter weights.
 
\begin{table}[h]
\centering
\caption{Per-block and total LoRA parameter count at rank $r=1$.}
\label{tab:lora-param-count}
\begin{tabular}{lrrr}
\toprule
Channel width $C$ & Params per block ($7C+768$) & Blocks & Total \\
\midrule
320  & 3{,}008 & 5  & 15{,}040 \\
640  & 5{,}248 & 5  & 26{,}240 \\
1280 & 9{,}728 & 6  & 58{,}368 \\
\midrule
\multicolumn{2}{r}{Total (16 blocks)} & 16 & \textbf{99{,}648} \\
\bottomrule
\end{tabular}
\end{table}

\begin{figure}[h]
\centering
\begin{subfigure}{.49\linewidth}
    \centering
    \includegraphics[width=\linewidth]{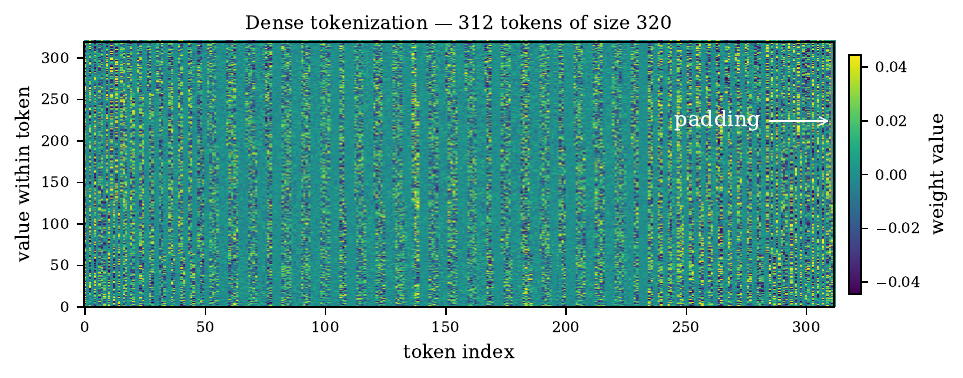}
    \caption{Dense tokenization. Only the final token is padded, marked in grey.}
    \label{fig:tokenization-dense}
\end{subfigure}
\begin{subfigure}{.49\linewidth}
    \centering
    \includegraphics[width=\linewidth]{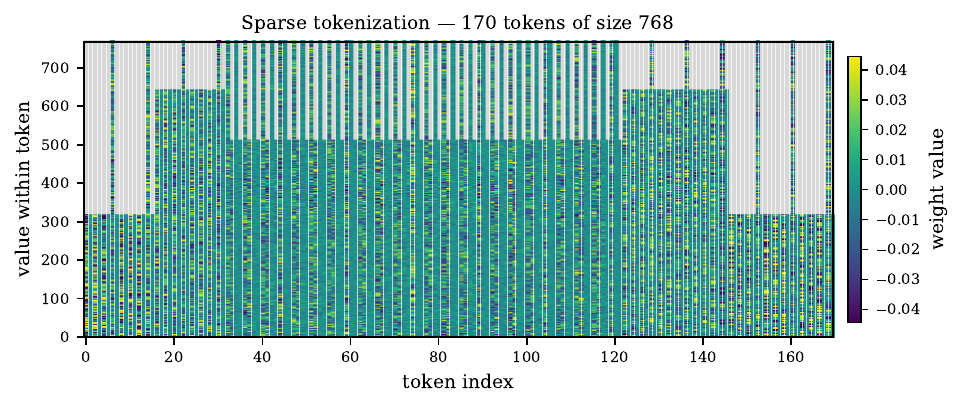}
    \caption{Sparse tokenization. The last token of each matrix is padded, marked in grey.}
    \label{fig:tokenization-sparse}
\end{subfigure}
\caption{Dense versus sparse tokenization of a single LoRA adapter.}
\label{fig:tokenization}
\end{figure}

\paragraph{Tokenization.} Dense tokenization (Figure~\ref{fig:tokenization-dense}) chunks the flattened adapter into fixed-size tokens of width $320$, the smallest channel width in our U-Net. The other channel widths are integer multiples of it, so every LoRA matrix except the cross-attention value (which operates on the $768$-dimensional text embedding) tiles into a whole number of tokens with no leftover. As an alternative, sparse tokenization (Figure~\ref{fig:tokenization-sparse}) slices each matrix into its own tokens, fully preserving channel boundaries at the cost of introducing a larger amount of padding. We use dense tokenization by default, since it already preserves the adapter's structure well enough while avoiding significant padding. For the mixed-rank setting, where rank-1 and rank-2 adapters must share one token sequence, we switch to the sparse tokenizer. Here, each matrix's token budget is fixed by its rank-2 size, so a rank-1 matrix occupies only the first half of that budget, with the remainder marked as padding by the mask.
 
\paragraph{Architecture.} Our primary autoencoder is a GPT-style, non-causal Transformer encoder-decoder following SANE~\citep{schurholt2024sane}. The encoder projects each of the $N=312$ input tokens from dimension $320$ to a model width of $1024$, applies $8$ layers with $16$ attention heads, and projects down to a latent dimension of $256$ per token. The decoder is symmetric.

\paragraph{Training.} The autoencoder is trained with a masked, batch-normalized MSE computed only over real-weight positions, for $1000$ epochs at batch size $32$ with AdamW~\citep{loshchilov2019adamw} (learning rate $10^{-4}$, weight decay $10^{-4}$) under a one-cycle schedule and a fixed seed of $2020$. We use dropout $0.15$, full precision, and gradient clipping at a max-norm of $10$. The adapter pool used for training is capped at $8{,}192$ adapters and split $70/15/15$ into train, validation, and test, giving $\sim\!5{,}734$ training adapters.

\clearpage
\section{Additional Experimental Results}

\subsection{Arithmetic and Interpolation}
\label{app:arithmetic-interpolation}

If the latent is organized, then simple vector operations on latent codes should correspond to meaningful, predictable changes in the decoded identity, which we refer to as ``face algebra.'' We test this directly by applying arithmetic and interpolation to the latents of real identities and inspecting the decoded results.

\paragraph{Setup.} Given two identities encoded to latents $z_A$ and $z_B$, we form new latents with a fixed set of elementary operations: the directional differences $z_A - z_B$ and $z_B - z_A$, the midpoint $(z_A + z_B)/2$, the raw sum $z_A + z_B$, the asymmetric combinations $z_A + z_B/2$ and $z_A + 2z_B$ that weight one identity more heavily, and a linear interpolation $z_\gamma = (1-\gamma)\,z_A + \gamma\,z_B$ between the two identities. For interpolation, we additionally apply the identical construction directly to the raw adapter weights~$\theta_A, \theta_B$.

\begin{figure}[H]
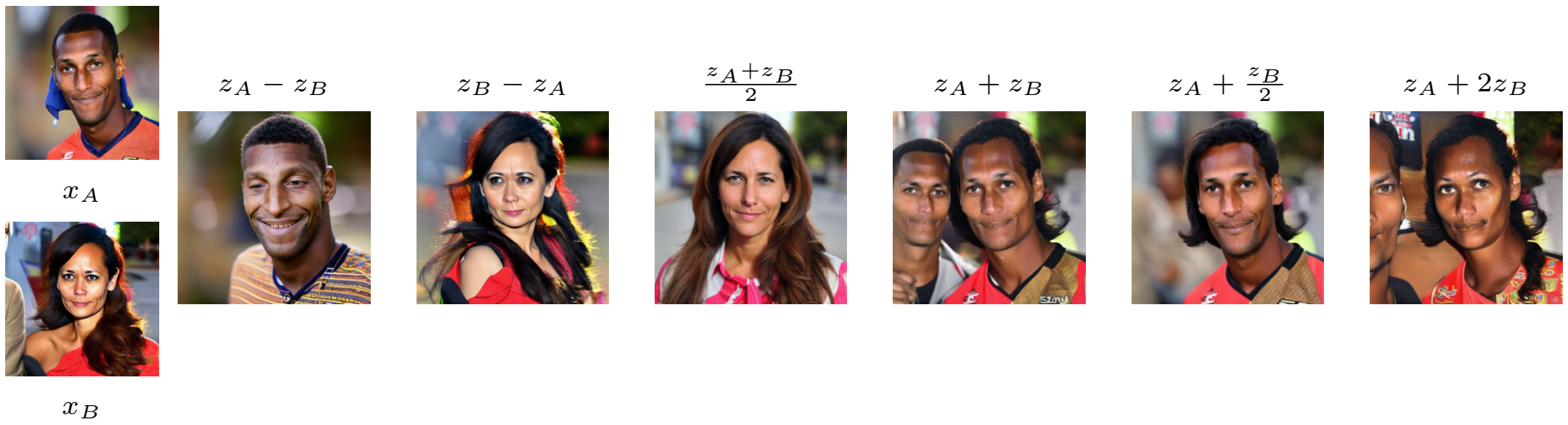

\centering

\newcommand{\inimg}[1]{%
    \includegraphics[width=0.10\linewidth]{#1}%
}

\newcommand{\opimg}[1]{%
    \includegraphics[width=0.125\linewidth]{#1}%
}

\newcommand{\vcentercell}[1]{%
    \ensuremath{\vcenter{\hbox{#1}}}%
}

\newcommand{\opcell}[2]{%
\raisebox{2.5mm}[0pt][0pt]{%
\vcentercell{%
\begin{tabular}{@{}c@{}}
{\scriptsize $#1$} \\[2pt]
\opimg{#2}
\end{tabular}%
}%
}%
}

\resizebox{\linewidth}{!}{%
\begin{tabular}{@{}c@{\hspace{5pt}}*{6}{c}@{}}

    \vcentercell{%
        \begin{tabular}{@{}c@{}}
            \inimg{figures/05\_arithmetic/latent\_pair\_4\_10/input\_a.png}
            \\[-1pt]
            {\scriptsize $x_A$}
            \\[3pt]
            \inimg{figures/05\_arithmetic/latent\_pair\_4\_10/input\_b.png}
            \\[-1pt]
            {\scriptsize $x_B$}
        \end{tabular}%
    }

    & \opcell{z_A-z_B}
        {figures/05\_arithmetic/latent\_pair\_4\_10/diff\_a\_minus\_b/sample\_0\_x\_hat.png}

    & \opcell{z_B-z_A}
        {figures/05\_arithmetic/latent\_pair\_4\_10/diff\_b\_minus\_a/sample\_0\_x\_hat.png}

    & \opcell{\frac{z_A+z_B}{2}}
        {figures/05\_arithmetic/latent\_pair\_4\_10/mean/sample\_0\_x\_hat.png}

    & \opcell{z_A+z_B}
        {figures/05\_arithmetic/latent\_pair\_4\_10/sum/sample\_0\_x\_hat.png}

    & \opcell{z_A+\frac{z_B}{2}}
        {figures/05\_arithmetic/latent\_pair\_4\_10/add\_scale0.5/sample\_0\_x\_hat.png}

    & \opcell{z_A+2z_B}
        {figures/05\_arithmetic/latent\_pair\_4\_10/add\_scale1.5/sample\_0\_x\_hat.png}

\end{tabular}%
}

\caption{
Arithmetic operations on the latents of two identities, $x_A$ and $x_B$, decode to visually sensible results, with the unscaled sum $z_A + z_B$ as the one exception that overloads the combination.
}
\label{fig:arithmetic-latent}
\end{figure}
\paragraph{Arithmetic works in the latent.} Elementary arithmetic on latent codes translates into visually sensible changes in the decoded identity (Figure~\ref{fig:arithmetic-latent}). For two encoded identities $A$ and $B$, the two directional differences $z_A - z_B$ and $z_B - z_A$ subtract one identity's features from the other in the expected direction. The midpoint $(z_A + z_B)/2$ decodes to a clean blend of the two people, and the weighted combinations tilt the result toward the more heavily weighted identity: $z_A + z_B/2$ gives a subtle influence of $B$ on $A$, whereas $z_A + 2z_B$ amplifies $B$ and pushes it into $A$. The raw sum $z_A + z_B$ overloads the combination, as expected of an operation that leaves the natural range of the latent. The same pattern holds for other identity pairs.

\begin{figure}[H]
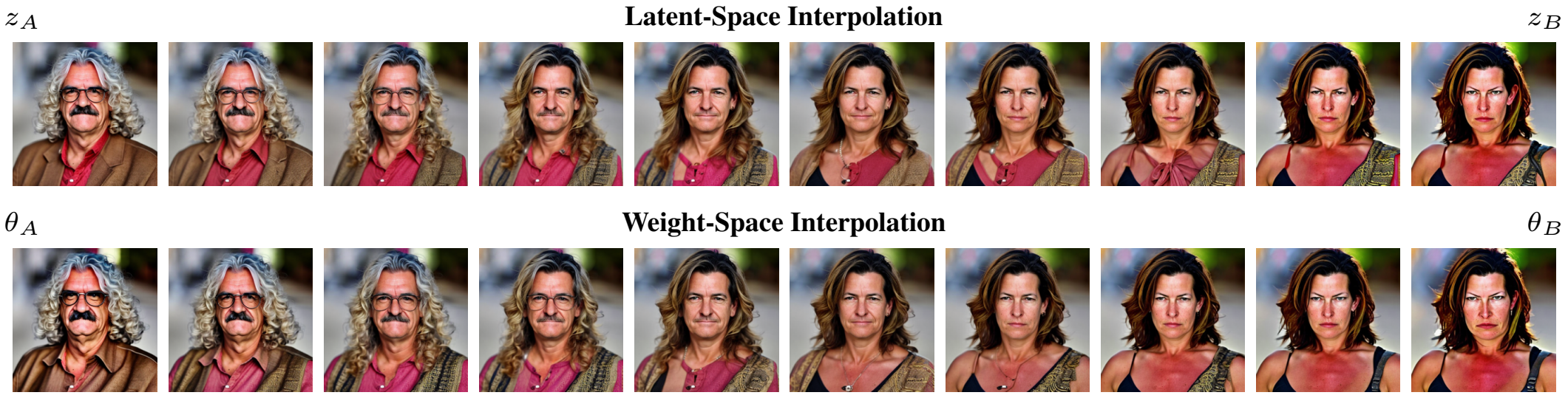

\centering

\setlength{\tabcolsep}{1.5pt}
\renewcommand{\arraystretch}{0.5}

\newcommand{\interpimg}[1]{%
    \includegraphics[
        width=0.092\linewidth,
        height=0.092\linewidth
    ]{#1}%
}

\newcommand{\interpheader}[3]{%
    \makebox[\linewidth][s]{%
        \scriptsize $#1$ \hfill \textbf{#2} \hfill $#3$%
    }\par\vspace{2pt}%
}

\interpheader{z_A}{Latent-Space Interpolation}{z_B}

\begin{tabular}{@{}cccccccccc@{}}
    \interpimg{figures/05\_interp/latents\_interpolated\_A3\_B7/sample\_0\_x\_hat.png} &
    \interpimg{figures/05\_interp/latents\_interpolated\_A3\_B7/sample\_1\_x\_hat.png} &
    \interpimg{figures/05\_interp/latents\_interpolated\_A3\_B7/sample\_2\_x\_hat.png} &
    \interpimg{figures/05\_interp/latents\_interpolated\_A3\_B7/sample\_3\_x\_hat.png} &
    \interpimg{figures/05\_interp/latents\_interpolated\_A3\_B7/sample\_4\_x\_hat.png} &
    \interpimg{figures/05\_interp/latents\_interpolated\_A3\_B7/sample\_5\_x\_hat.png} &
    \interpimg{figures/05\_interp/latents\_interpolated\_A3\_B7/sample\_6\_x\_hat.png} &
    \interpimg{figures/05\_interp/latents\_interpolated\_A3\_B7/sample\_7\_x\_hat.png} &
    \interpimg{figures/05\_interp/latents\_interpolated\_A3\_B7/sample\_8\_x\_hat.png} &
    \interpimg{figures/05\_interp/latents\_interpolated\_A3\_B7/sample\_9\_x\_hat.png}
\end{tabular}

\vspace{4pt}

\interpheader{\theta_A}{Weight-Space Interpolation}{\theta_B}

\begin{tabular}{@{}cccccccccc@{}}
    \interpimg{figures/05\_interp/input\_weight\_interpolation\_A3\_B7/sample\_0\_x\_hat.png} &
    \interpimg{figures/05\_interp/input\_weight\_interpolation\_A3\_B7/sample\_1\_x\_hat.png} &
    \interpimg{figures/05\_interp/input\_weight\_interpolation\_A3\_B7/sample\_2\_x\_hat.png} &
    \interpimg{figures/05\_interp/input\_weight\_interpolation\_A3\_B7/sample\_3\_x\_hat.png} &
    \interpimg{figures/05\_interp/input\_weight\_interpolation\_A3\_B7/sample\_4\_x\_hat.png} &
    \interpimg{figures/05\_interp/input\_weight\_interpolation\_A3\_B7/sample\_5\_x\_hat.png} &
    \interpimg{figures/05\_interp/input\_weight\_interpolation\_A3\_B7/sample\_6\_x\_hat.png} &
    \interpimg{figures/05\_interp/input\_weight\_interpolation\_A3\_B7/sample\_7\_x\_hat.png} &
    \interpimg{figures/05\_interp/input\_weight\_interpolation\_A3\_B7/sample\_8\_x\_hat.png} &
    \interpimg{figures/05\_interp/input\_weight\_interpolation\_A3\_B7/sample\_9\_x\_hat.png}
\end{tabular}
\caption{
Interpolating between two identities produces a smooth, valid transition whether performed in the latent (top) or directly in weight space (bottom).
}
\label{fig:interpolation-comparison}
\end{figure}
\paragraph{Interpolation also works in weight space.} Linearly interpolating between two encoded identities produces a smooth, gradual transition through valid intermediate identities (Figure~\ref{fig:interpolation-comparison}, top). Features change continuously from $A$ to $B$ with no abrupt switch and no invalid faces along the way. Applying the same construction directly to the raw weights (Figure~\ref{fig:interpolation-comparison}, bottom) yields a similarly recognizable transition between the two identities, showing that this operation is preserved even without the autoencoder.

\clearpage
\subsection{Editing}
\label{app:editing}

\begin{figure}[H]
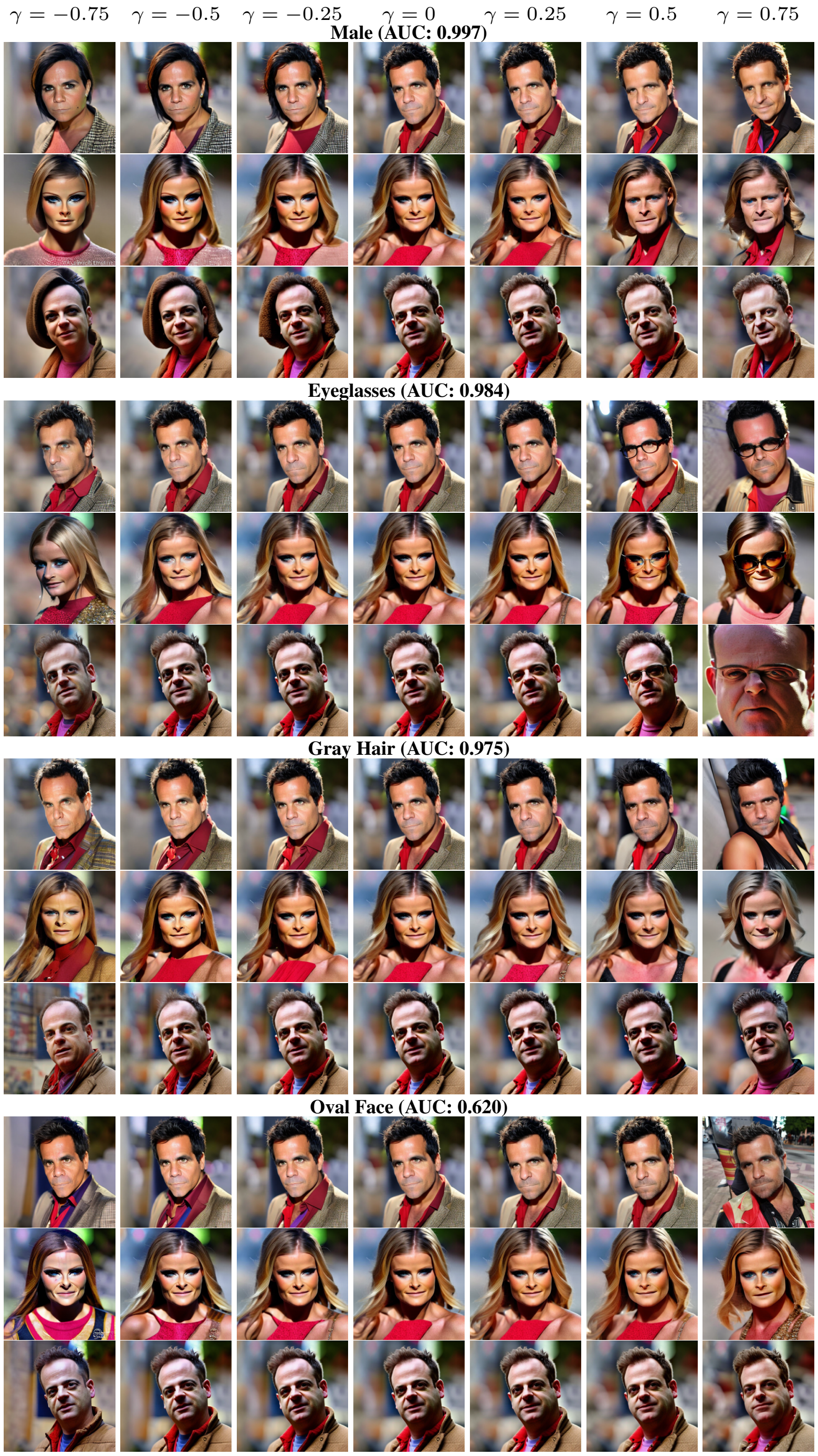

\centering

\setlength{\tabcolsep}{1pt}
\renewcommand{\arraystretch}{0.3}

\newcommand{\editimgappendix}[1]{%
    \includegraphics[width=0.1\linewidth,height=0.1\linewidth]{#1}%
}

\newcommand{\editrowappendix}[1]{%
    \editimgappendix{#1/sample\_0\_x\_hat.png} &
    \editimgappendix{#1/sample\_1\_x\_hat.png} &
    \editimgappendix{#1/sample\_2\_x\_hat.png} &
    \editimgappendix{#1/sample\_3\_x\_hat.png} &
    \editimgappendix{#1/sample\_4\_x\_hat.png} &
    \editimgappendix{#1/sample\_5\_x\_hat.png} &
    \editimgappendix{#1/sample\_6\_x\_hat.png}%
}

\begin{tabular}{@{}ccccccc@{}}
    \scriptsize $\gamma=-0.75$ &
    \scriptsize $\gamma=-0.5$ &
    \scriptsize $\gamma=-0.25$ &
    \scriptsize $\gamma=0$ &
    \scriptsize $\gamma=0.25$ &
    \scriptsize $\gamma=0.5$ &
    \scriptsize $\gamma=0.75$
    \\[1pt]

    \multicolumn{7}{c}{\scriptsize\textbf{Male  (AUC: 0.997)}} \\
    \editrowappendix{figures/06\_edit/edit\_Male\_sample2}
    \\[-0.5pt]
    \editrowappendix{figures/06\_edit/edit\_Male\_sample3}
    \\[-0.5pt]
    \editrowappendix{figures/06\_edit/edit\_Male\_sample5}
    \\[1pt]

    \multicolumn{7}{c}{\scriptsize\textbf{Eyeglasses  (AUC: 0.984)}} \\
    \editrowappendix{figures/06\_edit/edit\_Eyeglasses\_sample2}
    \\[-0.5pt]
    \editrowappendix{figures/06\_edit/edit\_Eyeglasses\_sample3}
    \\[-0.5pt]
    \editrowappendix{figures/06\_edit/edit\_Eyeglasses\_sample5}
    \\[1pt]

    \multicolumn{7}{c}{\scriptsize\textbf{Gray Hair (AUC: 0.975)}} \\
    \editrowappendix{figures/06\_edit/edit\_Gray\_Hair\_sample2}
    \\[-0.5pt]
    \editrowappendix{figures/06\_edit/edit\_Gray\_Hair\_sample3}
    \\[-0.5pt]
    \editrowappendix{figures/06\_edit/edit\_Gray\_Hair\_sample5}
    \\[1pt]

    \multicolumn{7}{c}{\scriptsize\textbf{Oval Face (AUC: 0.620)}} \\
    \editrowappendix{figures/06\_edit/edit\_Oval\_Face\_sample2}
    \\[-0.5pt]
    \editrowappendix{figures/06\_edit/edit\_Oval\_Face\_sample3}
    \\[-0.5pt]
    \editrowappendix{figures/06\_edit/edit\_Oval\_Face\_sample5}
\end{tabular}
\caption{
Additional identities for the four attributes in Figure~\ref{fig:steering-comparison}, showing that the effect is consistent across identities.
}
\label{fig:steering-comparison-extended}
\end{figure}



\end{document}